\documentclass[10pt,twocolumn,letterpaper]{article}

\usepackage{cvpr}              
\definecolor{cvprblue}{rgb}{0.21,0.49,0.74}
\usepackage[pagebackref,breaklinks,colorlinks,allcolors=cvprblue]{hyperref}
\usepackage[capitalize,noabbrev]{cleveref}
\usepackage{multirow}
\usepackage{pifont}
\usepackage{makecell}
\usepackage{bm}
\usepackage{amssymb}
\usepackage{threeparttable}
\usepackage{colortbl}
\usepackage[dvipsnames]{xcolor}
\usepackage[utf8]{inputenc}
\usepackage{diagbox}
\usepackage[accsupp]{axessibility}

\newcommand{\my}[1]{\textcolor{black}{#1}}

\newcommand{\appendixref}[1]{\hyperref[#1]{Appendix~\ref{#1}}}

\def\paperID{*****} 
\def\confName{CVPR}
\def\confYear{2026}

\title{Gaussian-Mixture Latent Flow for Stochastic 3D Human Motion Prediction}

\author{Yue Ma$^1$ \quad Frederick W. B. Li$^2$ \quad Xiaohui Liang$^{1,3,}$\thanks{Corresponding author} \\
$^1$Beihang University \quad $^2$Durham University \quad $^3$ Zhongguancun Laboratory\\
{\tt\small super\_mayue@buaa.edu.cn \quad frederick.li@durham.ac.uk \quad liang\_xiaohui@buaa.edu.cn}
}

\begin{document}
\maketitle

\begin{abstract}
Stochastic human motion prediction aims to forecast future motion distributions. Although recent studies have achieved strong performance in terms of accuracy and diversity, they often overlook plausibility (e.g., resulting in physically unrealistic predictions) and uncertainty quantification, both of which are essential for real-world applications and downstream tasks. To address these issues, we propose a latent flow-based model equipped with a data-driven Gaussian mixture prior that more effectively disentangles diverse human behaviors than conventional single-modal priors. This prior is derived from patterns in the training data without requiring additional annotations. Furthermore, the fully invertible nature of our model enables natural uncertainty quantification through tractable likelihood computation. Experiments on the Human3.6M and AMASS datasets demonstrate that our approach achieves state-of-the-art performance in both accuracy and plausibility. 
\end{abstract}

\section{Introduction}
Human motion prediction (HMP) aims to forecast future 3D human poses from observed human motion.
As a fundamental problem in computer vision and robotics, HMP has become essential to a wide range of real-world applications and downstream tasks, including autonomous driving ~\cite{hu2023planning,casas2021mp3,chen2024end}, human robot collaboration (HRC) ~\cite{liu2017human,zhao2020experimental,li2022proactive,kanazawa2019adaptive}, and assistive robotics~\cite{lee2022robot,teramae2017emg}. In these contexts, AI-driven agents rely on HMP approaches to anticipate human behavior in dynamic environments, thereby enabling proactive motion planning, collision avoidance, and more natural human robot interaction. Since many previous studies ~\cite{martinez2017human,li2020dynamic,mao2020history,dai2023kd,sun2023defeenet,gao2023decompose,guo2023back} formulated HMP as a regression problem predicting a single future motion while neglecting its inherently multi-modal nature, recent stochastic motion prediction methods ~\cite{barsoum2018hp,g2017deligan,gui2018adversarial,yuan2019diverse,yuan2020dlow,mao2021GSPS,barquero2023belfusion,wei2023MotionDiff,chen2023humanmac,sun2024comusion} have instead focused on learning a probability distribution over possible future motions. Consequently, the objective of HMP extends beyond mere \textit{accuracy} to encompass \textit{diversity} and \textit{plausibility}, 
while faithfully modeling the predictive uncertainty of each future trajectory.

\begin{figure}[tbp]
    \centering
    \begin{tabular}{ll}
    \subfloat[multi-modal flow-based method (ours)]{
        \includegraphics[width=0.9\linewidth]{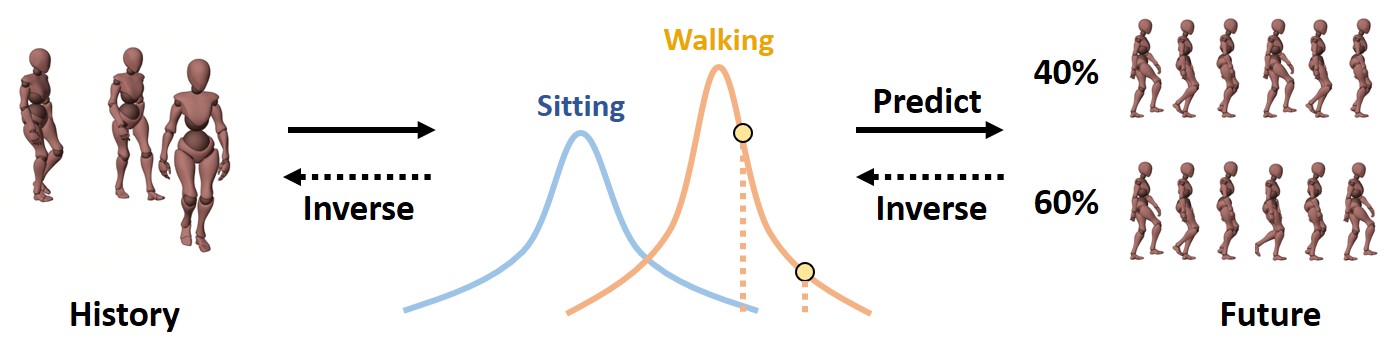}    
        \label{fig:teaser_1}
    } \\
    \subfloat[single-modal generative model-based methods]{
        \includegraphics[width=0.9\linewidth]{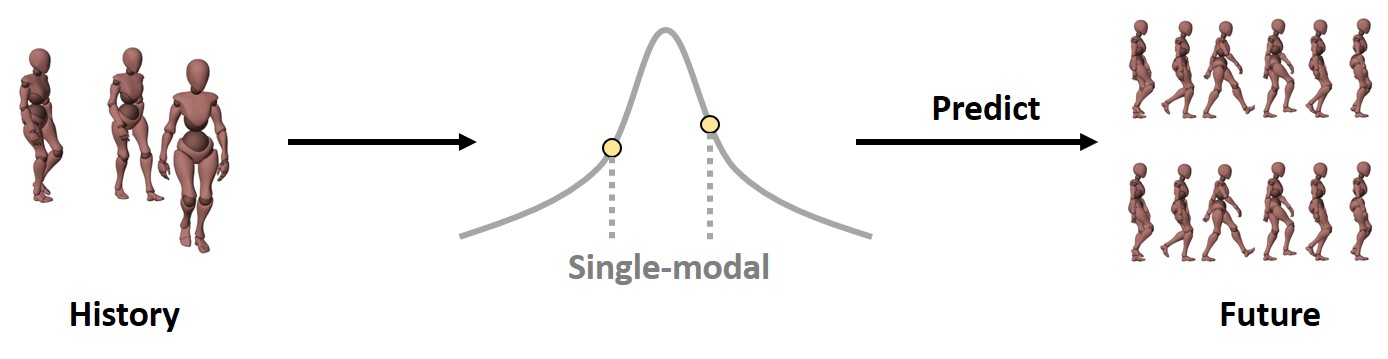}    
        \label{fig:teaser_2}
    }
    \end{tabular}
    \caption{Unlike prior approaches, our method (1) introduces a mixed Gaussian prior to effectively disentangle diverse human motion patterns, improving \textit{plausibility} by reducing semantic entanglement, and (2) incorporates a fully invertible architecture that supports exact likelihood computation, thereby providing a principled means of uncertainty estimation.}
    \label{fig:teaser}
\end{figure}

Despite progress with deep generative models, e.g., variational autoencoders (VAEs) ~\cite{kingma2013auto}
and diffusion models~\cite{ho2020denoising}, several critical challenges still persist in stochastic HMP: 1) The sacrifice of \textit{plausibility} for \textit{accuracy} and \textit{diversity}, leading to physically infeasible or physiologically unnatural predictions (e.g., anatomically impossible joint angles); 2) Insufficient modeling and quantification of \textit{uncertainty}. The challenge about \textit{plausibility} primarily originates from imposing a single-modal prior (e.g., the standard Gaussian prior in VAE-based methods~\cite{yuan2020dlow,
mao2021GSPS,dang2022diverse,xu2022diverse}) 
on multi-modal human behavior patterns (e.g., walking vs. sitting), which ultimately result in predictions that conflate distinct behavioral modes. The \textit{uncertainty} limitation in most prior works \cite{yuan2020dlow,mao2021GSPS,sun2024comusion,curreli2025nonisotropic} derives the likelihood approximation during distribution forecasting (e.g., the variational lower bound (VLB) used in VAEs and diffusion models), which fails to serve as a reliable measure of uncertainty, and thereby cannot effectively balance multi-modal prediction with accurate uncertainty estimation.

Inspired by recent advances in normalizing flows~\cite{dinh2014nice,dinh2016density,kingma2018glow} and flow matching~\cite{liu2022flow,lipman2023flow,kim2024simulation}, we propose a flow-based approach to address these challenges, as shown in \cref{fig:teaser}. Similar to state-of-the-art latent diffusion-based methods~\cite{barquero2023belfusion,sun2024comusion,curreli2025nonisotropic}, our approach also operates in the latent space. However, we introduce a data-driven multi-modal prior, represented as a mixture of Gaussians, to adaptively model diverse motion patterns in the latent space. This formulation effectively mitigates semantic confusion in forecasting and thus enhances \textit{plausibility}. Moreover, we adopt the flow model~\cite{ma2025probHMI} as the latent backbone in place of the VAE~\cite{barquero2023belfusion} or autoencoder (AE)~\cite{sun2024comusion,curreli2025nonisotropic}, and replace the SDE-based training of diffusion models with an ODE-based flow-matching framework. This design preserves model invertibility and enables tractable likelihood computation for natural uncertainty quantification. Since flow matching approaches eliminate the data-space supervision commonly employed by latent diffusion-based methods~\cite{barquero2023belfusion,sun2024comusion,curreli2025nonisotropic} for improving \textit{accuracy}, we additionally introduce a skeleton-aware transformer to model the latent velocity field, incorporating skeletal priors to enhance both \textit{accuracy} and \textit{plausibility}. In summary, our main contributions are as follows:
\begin{itemize}
\item We propose the first latent flow matching-based method for HMP, which incorporates a skeleton-aware transformer and is fully invertible. 
\item To overcome the limitations of single-modal priors, we propose a multi-modal Gaussian prior for the latent space. Given the scarcity and unreliability of motion sequence labels, we utilize an Expectation Maximization algorithm (EM) for constructing the mixed distribution in an unsupervised manner.
\item We evaluate our method against SOTA approaches on two large-scale MoCap datasets: Human3.6M~\cite{ionescu2013human3} and AMASS~\cite{AMASS2019}. The experimental results demonstrate superior \textit{accuracy} and \textit{plausibility} of our method. 
\end{itemize}

\section{Related Work}

\subsection{3D Human Motion Prediction}
Deterministic prediction approaches~\cite{pavllo2020modeling,mao2020history,li2020dynamic,li2021multiscale,zhong2022spatio,sun2023defeenet,gao2023decompose,nargund2023spotr,wang2023graph,guo2023back,chen2023mstp} treat HMP as a regression task, aiming to predict a single most likely motion sequence. Since these methods ignore the multi-modal nature inherent in human motion, stochastic prediction methods leverage deep generative models to forecast the distribution of future motion, including generative adversarial networks (GANs) \cite{barsoum2018hp,g2017deligan,gui2018adversarial}, VAEs \cite{b2018accurate,yuan2019diverse,zhang2021we,ma2022multi,yuan2020dlow,mao2021GSPS,dang2022diverse,salzmann2022motron} and diffusion models \cite{barquero2023belfusion,wei2023MotionDiff,chen2023humanmac,sun2024comusion,xu2024learning,tian2024transfusion,curreli2025nonisotropic}. These approaches generally operate by mapping a simple prior distribution to the distribution of future motions, conditioned on the observed motion. However, this framework confronts two challenges. First, the imposition of a single-modal prior on multi-modal motions causes existing methods to conflate distinct behaviors. Although some studies ~\cite{yuan2020dlow,xu2022diverse,dang2022diverse} attempt to transform samples towards less-likely areas of the latent space, their foundation in single-modal assumptions persists. Second, the intractable likelihood inherent to these methods poses a significant barrier to uncertainty estimation, thereby limiting the utility in safety-aware scenarios.

Several works have studied \textit{uncertainty} in HMP. UA-HMP~\cite{ding2021uncertainty} and SAGGB~\cite{wang2024existence} use variance as the uncertainty measure but rely on simplistic Gaussian assumptions. Saadatnejad et al.~\cite{saadatnejad2024toward} regress prediction error with an exponential distribution, also imposing restrictive assumptions. Additionally, these methods are limited to deterministic prediction. In stochastic settings, Motron~\cite{salzmann2022motron} employs a Gaussian distribution on $\mathrm{SO}(3)$ as a parametric output structure for sampling diverse motions and computing likelihoods, but it lacks constraints needed to ensure realistic predictions. DE-TGN~\cite{eltouny2024tgn} employs model ensembles to capture uncertainty but incurs high computational cost. ProbHMI \cite{ma2025probHMI} employs invertible networks to map motions into a latent Gaussian space, yet its uncertainty remains approximate due to quantile-based estimation in the latent space.

\subsection{Probabilistic Flow Models}
Normalizing flows~\cite{dinh2014nice,dinh2016density,kingma2018glow,behrmann2019invertible,chen2019residual} construct complex probability distributions by applying a series of invertible transformations to a tractable base distribution. Due to the constraints of maintaining invertibility and computing the Jacobian within each layer, continuous normalizing flows (CNFs)~\cite{chen2018neural,mathieu2020riemannian,bilovs2021neural} extend discrete transformations into continuous ones, representing the invertible process as an ordinary differential equation (ODE). Although CNFs are highly expressive, their training is computationally expensive due to the need for ODE simulations at each iteration. To address this issue, flow matching~\cite{lipman2022flow,abdal2021styleflow,liu2022flow,albergo2022building,lipman2023flow,kim2024simulation} provides a simulation-free training paradigm for CNFs, significantly reducing computational cost.

Although most flow-based approaches adopt a standard Gaussian as the base distribution, such as FloMo~\cite{scholler2021flomo} and FlowChain~\cite{maeda2023fast} for 2D trajectory prediction, and MoGlow~\cite{henter2020moglow} and its variant~\cite{yin2021graph} for motion generation, several studies have explored mixture-based alternatives. FlowGMM~\cite{izmailov2020semi} and STG-NF~\cite{hirschorn2023normalizing} employ Gaussian mixture models as priors for classification and anomaly detection tasks respectively, and StyleVR~\cite{ji2023stylevr} utilizes a Student’s t mixture model for stylized motion generation. However, all of these methods rely on auxiliary labels to construct mixture distributions. MGF \cite{chen2024mgf} uses an unsupervised mixed prior to represent diverse motion patterns; however, their method predefines the mixture via clustering on 2D trajectories, which is ineffective for 3D human motion due to its higher dimensionality and complex spatial structure. In contrast, our EM-based strategy learns the mixture distribution directly during training, thereby more effectively capturing meaningful motion semantics in the latent space.

\section{\my{Preliminary}}
Flow matching formulates the objective of CNFs through a velocity field $v$, which defines a probability path between two distributions, $\pi_{0}$ and $\pi_{1}$, as follows:
\begin{equation}
    \mathrm{d}\mathbf{x}_{t} = v(\mathbf{x}_{t}, t)\mathrm{d}t, \quad t \in [0, 1],
    \label{EQ:ode}
\end{equation}
\noindent where $\mathbf{x}_{1} \sim \pi_{1}$ and $\mathbf{x}_{0} \sim \pi_{0}$. The transformation from $\pi_{0}$ and $\pi_{1}$ can then be achieved by integrating \cref{EQ:ode}. Since flow matching optimizes only the velocity field $v$, it avoids the time-consuming ODE integration required for explicit likelihood computation during training.

Rectified Flow~\cite{liu2022flow} provides an effective formulation for modeling the velocity field $v$. Specifically, it defines $v$ as a linear interpolation between $\mathbf{x}_{1}$ and $\mathbf{x}_{0}$, yielding
\begin{equation}
    \mathbf{x}_{t} = t\mathbf{x}_{1} + (1{-}t)\mathbf{x}_{0},
\end{equation}
where $\mathbf{x}_{t}$ evolves along the straight-line direction $(\mathbf{x}_{1} - \mathbf{x}_{0})$ with constant velocity. Accordingly, the objective of the velocity field $v$ in Rectified Flow can be formulated as
\begin{equation}
    \min_{\substack{v}} \int_{0}^{1}     \mathbb{E}\big[\Vert (\mathbf{x}_{1} - \mathbf{x_{0}}) - v(\mathbf{x}_{t}, t) \Vert_{2}^{2}\big] \mathrm{d}t.
\end{equation}

\section{Problem Formulation} \label{SEC:problem_formulation}
We aim to predict a diverse set of possible future 3D human motions, along with their associated probability densities, based on past movements. Formally, given a motion sequence $\mathbf{X} = (\mathbf{x}_1, \mathbf{x}_2, \cdots, \mathbf{x}_{T+N}) \in \mathbb{R}^{(T+N) \times J \times C}$, the observed sequence is defined as the first $T$ frames, $\mathbf{X}_{obs} = (\mathbf{x}_1, \mathbf{x}_2, \cdots, \mathbf{x}_{T})$, and the future sequence as the subsequent $N$ frames, $(\mathbf{x}_{T+1}, \mathbf{x}_{T+2}, \cdots, \mathbf{x}_{T+N})$. Here, $J$ and $C$ denote the number of joints and the number of channels per joint, respectively. For clarity, we denote the predicted future motion as $\hat{\mathbf{Y}}$ and the corresponding ground-truth future as $\mathbf{Y}$, while $\hat{\mathbf{X}}$ represents the reconstructed full sequence corresponding to $\mathbf{X}$. The associated log-probabilities, which naturally quantify uncertainty, are denoted as $\log p(\hat{\mathbf{X}})$. A set of $M$ diverse predictions is then denoted as $\{\hat{\mathbf{Y}}^1, \hat{\mathbf{Y}}^2, \ldots, \hat{\mathbf{Y}}^M\}$.

\section{Methodology}

\begin{figure*}[htb]
    \centering
    \includegraphics[width=0.8\linewidth]{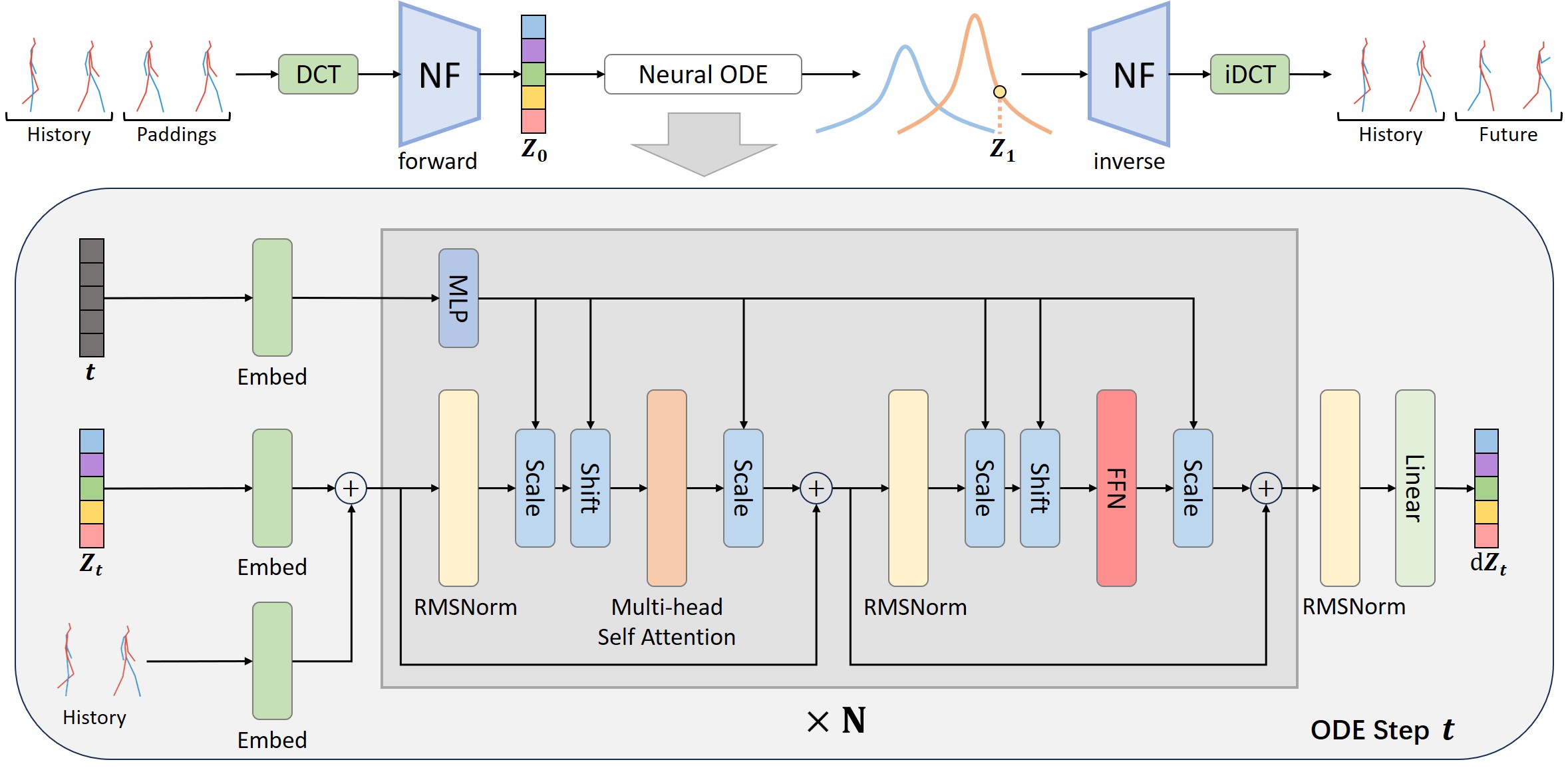}
    \caption{
    \textbf{Overview of the framework.} We predict future motion within a latent space constructed by a flow model. Owing to the invertibility of normalizing flows, a single model can function as both an encoder and a decoder through its forward and inverse processes, respectively. During latent forecasting, we first pad the observed sequence to the full length using the final observed frame and transform it into the latent space to obtain the starting point $\mathbf{Z}_{0}$. In training, the model learns the velocity field that maps $p_{\mathbf{Z}_{0}} = \mathcal{N}(\mathbf{Z}_{0}, \mathrm{I})$ to $\mathbf{Z}_{1}$, corresponding to the ground-truth future. In inference, we employ an ODE solver to evolve the latent state over time, and the resulting latent codes are decoded to generate the final motion predictions.
    }
    \label{fig:overview}
\end{figure*}


Human motion exhibits complex and multi-modal patterns that are poorly captured by the single-modal prior used in prior works~\cite{barquero2023belfusion,curreli2025nonisotropic}, leading to semantic entanglement and reduced plausibility in predictions. To overcome these challenges, we propose a two-stage framework for stochastic HMP that advances prior multi-modal approaches such as~\cite{izmailov2020semi}. First, we utilize an unsupervised EM algorithm to learn a Gaussian mixture model as the latent prior, inherently disentangling distinct motion behaviors without auxiliary labels, as detailed in \cref{SEC:latent_space}. Second, we model latent dynamics via a neural ODE that ensures full invertibility and enables exact likelihood computation for principled uncertainty quantification, as described in \cref{SEC:latent_flow_model,SEC:inference}. 



\subsection{Gaussian-Mixture Latent Representation} \label{SEC:latent_space}
We utilize a part-aware flow model~\cite{ma2025probHMI} as the latent backbone for encoding and decoding motion sequences. By applying transformations over hierarchical body parts, the model preserves the semantic structure of each joint in the latent space, making it well aligned with the requirements of our skeleton-aware transformer (see \cref{SEC:latent_flow_model}). Similar to previous works~\cite{mao2021GSPS,xu2022diverse,chen2023humanmac}, we first project the sequence into the frequency domain using the Discrete Cosine Transform (DCT) and apply a low-pass filter to smooth the sequence, which promotes the generation of continuous motions. Specifically, given $\mathbf{X} \in \mathbb{R}^{(T+N) \times J \times C}$, we first flatten the joint and channel dimensions into a single spatial axis, represented as $\mathbf{X} \in \mathbb{R}^{(T+N) \times JC}$, and then apply a 2D-DCT to compress the data along both the two dimensions, with a 2D-iDCT for reconstruction, formulated as:
\begin{equation}
    \begin{split}
    \tilde{\mathbf{X}} &= {\rm DCT}(\mathbf{X}) = D_{L_{1}} \mathbf{X}G_{L_{2}} \\
    \mathbf{X} &= {\rm iDCT}(\tilde{\mathbf{X}}) = D^{T}_{L_{1}}\tilde{\mathbf{X}}G^{T}_{L_{2}},    
    \end{split}
    \label{EQ:DCT}
\end{equation}
\noindent where 2D-iDCT $D_{L_{1}} \in \mathbb{R}^{L_{1} \times (T+N)}$ and $G_{L_{2}} \in \mathbb{R}^{C \times L_{2}}$ denote the predefined DCT basis matrices, retaining the first $L_{1}$ and $L_{2}$ low-frequency components for the temporal and channel dimensions, respectively. In practice, we set $L_{2} = C$, resulting in $\tilde{\mathbf{X}} \in \mathbb{R}^{L_{1} \times JC}$. For brevity, we refer to $L_{1}$ simply as $L$ in the following sections, denote $\tilde{\mathbf{X}}$ as $\mathbf{X}$ whenever the context is clear.

\vspace{-0.3cm}
\paragraph{Latent Space Construction.}
We transform the data distribution into a mixed distribution comprising $K$ Gaussian components, denoted as $\{\mathcal{N}(\bm{\mu}_{i}, \bm{\sigma}_{i}), i{=}1 \dots K\}$. The distributional parameters are estimated by fitting the model to the training data, enabling the flow model to effectively disentangle and represent multiple motion patterns within the dataset. Specifically, we employ EM algorithm to jointly optimize the parameters of both the flow model and the mixture distribution. During the E-step, we compute posterior probabilities $q(i|\mathbf{Z})$ of latent variables for each motion sequence, formulated as:
\begin{align}
    q(i|\mathbf{Z}) = \frac{p(i, \mathbf{Z})} {p(\mathbf{Z})} &= \frac{\beta_i\mathcal{N}\big(f_{\theta}(\mathbf{X})|\bm{\mu}_{i}, \bm{\sigma}_{i}\big) |\det(\frac{\partial{f_{\theta}}}{\partial{\mathbf{X}}})|}{\sum_{j=1}^{K} \beta_j\mathcal{N}\big(f_{\theta}(\mathbf{X})|\bm{\mu}_{j}, \bm{\sigma}_{j}\big) |\det(\frac{\partial{f_{\theta}}}{\partial{\mathbf{X}}})|} \notag \\
    &= \frac{\beta_i\mathcal{N}\big(f_{\theta}(\mathbf{X})|\bm{\mu}_{i}, \bm{\sigma}_{i}\big)}{\sum_{j=1}^{K} \beta_j\mathcal{N}\big(f_{\theta}(\mathbf{X})|\bm{\mu}_{j}, \bm{\sigma}_{j}\big)},
    \label{EQ:e_step}
\end{align}
\noindent where $\mathbf{Z} {=} f_{\theta}(\mathbf{X})$ is the latent variable, $f_{\theta}$ represents the flow model, and $\beta_{i}$ denotes the mixture coefficient.

During the M-step, we first update $\beta_{i}$ according to \cref{EQ:e_step}, following the standard procedure used in conventional GMMs. Subsequently, we update the parameters of each sub-distribution $(\bm{\mu}_{i}, \bm{\sigma}_{i})$ with the fixed $\theta$, using a soft assignment strategy, which corresponds to maximizing the objective defined in 
\cref{EQ:distribution_update}:
\begin{equation}
    \mathbb{E}_{p_{\rm{data}}}\Big[ {q(i|\mathbf{Z})}\log\Big(\mathcal{N}\big(f_{\theta}(\mathbf{X})|\bm{\mu}_{i}, \bm{\sigma}_{i}\big)|\det(\frac{\partial{f_{\theta}}}{\partial{\mathbf{X}}})|\Big)\Big].    
    \label{EQ:distribution_update}
\end{equation}

Next, we update the flow model parameters $\theta$ using the updated mixture distribution, employing a hard assignment that maximizes the likelihood of each $\mathbf{X}$ under its corresponding Gaussian component. This hard assignment strategy prevents mode collapse during training, which could otherwise result in a single-modal latent distribution. The optimization objective can then be formulated as:
\begin{equation}    
    \sum_{i=1}^K\Big[\mathbb{I}(i) \log\Big(\mathcal{N}\big(f_{\theta}(\mathbf{X})|\bm{\mu}_{i}^{new}, \bm{\sigma}_{i}^{new} \big)|\det(\frac{\partial{f_{\theta}}}{\partial{\mathbf{X}}})|\Big)\Big],
    \label{EQ:flow_model_update}
\end{equation}
\noindent where $\mathbb{I}(\cdot)$ denotes the indicator function, such that $\mathbb{I}(i){=}1$ if and only if the likelihood of $\mathbf{X}$ under the $j$-th Gaussian component is maximal.

By iteratively performing the E-step and M-step, we obtain the latent space corresponding to a data-driven Gaussian mixture distribution $q_{z}(\mathbf{Z})$ defined as:
\begin{equation}    
    q_{z}(\mathbf{Z}) = \sum_{i=1}^{K} \beta_{i}\mathcal{N}(\bm{\mu}_{i}, \bm{\sigma}_{i}).
    \label{EQ:mixture_distrib}
\end{equation}

\subsection{Latent Flow Matching} \label{SEC:latent_flow_model}
We predict future motion within the constructed latent space. Given the observation $\mathbf{X}_{obs}$, we first pad it with the final observed frame to form a complete sequence before encoding, obtaining the latent representation $\mathbf{Z}_{0}$. The latent code is then evolved toward the ground-truth latent code $\mathbf{Z}_{1}$ through an ODE formulation. To enable stochastic prediction, we reframe it as a transport between two distributions, $p_{\mathbf{Z}_{0}}$ and $p_{\mathbf{Z}_{1}}$, where $p_{\mathbf{Z}_{0}} = \mathcal{N}(\mathbf{Z}_{0}, \mathrm{I})$ and $p_{\mathbf{Z}_{1}}$ is modeled as a Dirac distribution $\delta(\mathbf{Z_{1}})$. By defining the conditional probability path as a linear interpolation~\cite{liu2022flow} between $p_{\mathbf{Z}_{0}}$ and $p_{\mathbf{Z}_{1}}$, the objective of the velocity field $v_{\theta}$ is defined as:
\begin{equation}
    \mathbb{E}_{\mathbf{Z}_{1} \sim p_{\rm{data}}, \mathbf{\hat{Z}}_{0} \sim p_{\mathbf{Z_{0}}}, t \sim [0, 1]} \Vert (\mathbf{Z}_{1} - \mathbf{\hat{Z}}_{0}) - v_{\theta}(\mathbf{Z}_{t}, t) \Vert_{2}^{2},
    \label{EQ:fm_objective}
\end{equation}
\noindent where $\mathbf{Z}_{t} = t\mathbf{Z}_{1} + (1{-}t)\mathbf{\hat{Z}}_{0}$.


We implement $v_{\theta}$ with a skeleton-aware transformer model, as illustrated in \cref{fig:overview}. Specifically, our model uniquely tokenizes temporal trajectories of individual joints rather than poses, enabling 1) joint-wise attention that explicitly models spatial dependencies within the human skeleton, 2) temporal embeddings that facilitate effective alignment and conditioning on observed motion history. These components jointly preserve anatomical consistency and temporal context, which are key factors often overlooked in earlier transformer-based methods~\cite{chen2023humanmac,barquero2023belfusion,sun2024comusion}.

\vspace{-0.3cm}
\paragraph{Input Embedding.}
We align $\mathbf{Z}_t \in \mathbb{R}^{L \times JC}$ and the observation $\mathbf{X}_{obs} \in \mathbb{R}^{T \times JC}$ to a unified temporal dimension through linear embeddings, obtaining $\mathbf{Z}_t^{'} \in \mathbb{R}^{L_{out} \times JC}$ and $\mathbf{X}_{obs}^{'} \in \mathbb{R}^{L_{out} \times JC}$, which are then fused through element-wise addition. The temporal alignment enables direct integration of the observed context as the condition signal, thereby strengthening the historical-future dependency. This contrasts with prior methods~\cite{chen2023humanmac,barquero2023belfusion,sun2024comusion} that typically formulate the task as a reconstruction problem and fail to explicitly capture this essential relational structure. 

\vspace{-0.3cm}
\paragraph{Feature-wise Modulation.}
Inspired by successful applications of transformers in diffusion-based image generation~\cite{peebles2023scalable,yao2025reconstruction}, we modulate latent features conditioned on flow matching time step $t$. Specifically, we apply an affine transformation after each RMSNorm~\cite{zhang2019root} layer and introduce a scaling transformation immediately before each residual connection within the transformer block. All parameters of modulation are generated by an MLP module~\cite{dilokthanakul2016deep}, which takes the embedding vector of $t$ as input.

\vspace{-0.3cm}
\paragraph{Joint-wise Attention.}
We compute the scaled dot-product self-attention~\cite{vaswani2017attention} within the multi-head attention along the $J$ and $C$ dimension, as follows: 
\begin{equation}
    {\rm Attention}(\mathbf{Q}_i, \mathbf{K}_i, \mathbf{V}_i) = {\rm softmax}(\frac{\mathbf{Q}_i^{T}\mathbf{K}_i}{\sqrt{d_k}})\mathbf{V}_i,
\end{equation}
\noindent where $\mathbf{Q}_i, \mathbf{K}_i, \mathbf{V}_i$ are the query, key, and value matrices for the $i$-th attention head, and $d_k$ is a scaling factor. Given the input feature $\mathbf{F} \in \mathbb{R}^{L_{out} \times JC}$, $\mathbf{Q}_i, \mathbf{K}_i, \mathbf{V}_i \in \mathbb{R}^{L_{out} \times JC}$ are defined as:
\begin{equation}
    \mathbf{Q}_i, \mathbf{K}_i = {\rm RMSNorm}\big(g(\mathbf{F})\big), 
    \mathbf{V}_i = g(\mathbf{F}).
\end{equation}

Here, $g: \mathbb{R}^{L_{out}} \to \mathbb{R}^{L_{out}}$ denotes a Linear layer, and is also applied within the feed-forward network (FFN). 

\subsection{Inference and Likelihood Computation} \label{SEC:inference}
During inference, we generate the full motion sequence $\mathbf{\hat{X}}$ and its corresponding log-likelihood $\log p(\mathbf{\hat{X}})$, serving as an uncertainty measure, from an initial latent sample $\mathbf{\hat{Z}_{0}} \sim \mathcal{N}(\mathbf{Z_{0}}, \mathrm{I})$ through a two-stage likelihood-tractable process. In the first stage, we integrate the ODE defined by $v_{\theta}(\mathbf{Z}_t, t)$, starting from $\mathbf{\hat{Z}_{0}}$, to obtain the predicted latent variable $\mathbf{\hat{Z}_{1}}$. The log-likelihood of $\mathbf{\hat{Z}_{1}}$ is then computed as:
\begin{equation}
    \log p(\mathbf{\hat{Z}_{1}}) = \log p(\mathbf{\hat{Z}_{0}}) + \int_{0}^{1} - \mathrm{tr}\Big(\nabla_{\mathbf{Z}_t}v_{\theta}(\mathbf{Z}_t, t)\Big) \mathrm{d}t,
    \label{EQ:likelihood_stage_1}
\end{equation}
\noindent where $\mathrm{tr}(\cdot)$ is estimated using the Hutchinson’s trace estimator~\cite{grathwohl2018ffjord, chen2018neural} in practice.

In the second stage, the latent variable $\mathbf{\hat{Z}_{1}}$ is mapped back to the motion space to obtain $\mathbf{\hat{X}}$ through the reverse process of the flow model. Consequently, the log-likelihood of $\mathbf{\hat{X}}$ can be expressed as:
\begin{align}
    \label{EQ:likelihood} 
    \log p(\mathbf{\hat{X}}) &= \log p(\mathbf{\hat{Z}_{1}}) + \log{|\det(\frac{\partial{\mathbf{\hat{X}}}}{\partial{\mathbf{\hat{Z}_{1}}}})|} \\
    &= \log p(\mathbf{\hat{Z}_{0}}) + \int_{0}^{1} - \mathrm{tr}(\cdot) \mathrm{d}t + \log{|\det(\frac{\partial{\mathbf{\hat{X}}}}{\partial{\mathbf{\hat{Z}_{1}}}})|}. \notag
\end{align}

While the log-likelihood of $\mathbf{\hat{X}}$ can be directly computed using the mixed Gaussian distribution defined in \cref{EQ:mixture_distrib}, it only reflects the global data distribution. In contrast, the conditional likelihood given the observed history, defined in \cref{EQ:likelihood}, provides a more faithful uncertainty measure, as it evaluates the likelihood that the predicted motion occurs under the current scenario, thereby supporting safety-aware decision making in downstream tasks to priority responses.

\section{Experiments}\label{SEC:Experiments}
\subsection{Experimental Setups}
\paragraph{Datasets.}
We evaluate our approach using Human3.6M \cite{ionescu2013human3} and AMASS \cite{AMASS2019} datasets. For Human3.6M, we set the numbers of observation and prediction frames to 25 (0.5 s) and 100 (2 s), respectively. Following DLow~\cite{yuan2020dlow}, we adopt a 17-joint skeleton and use subjects S1, S5, S6, S7, and S8 for training, while S9 and S11 are reserved for testing. For AMASS, we set the numbers of observation and prediction frames to 30 (0.5 s) and 120 (2 s), respectively. Following BeLFusion~\cite{barquero2023belfusion}, we employ a 22-joint skeleton, using eleven sub-datasets for training and four sub-datasets for testing. All motion poses are represented using exponential maps to ensure a consistent skeletal parameterization.

\begin{table*}[htb]
    \renewcommand{\arraystretch}{1.25}
    \centering
    \caption{Quantitative results compared to stochastic baselines adopting \textbf{Best-of-50} metrics on the Human3.6M and AMASS datasets. As AMASS does not include action labels, FID is not used for evaluation. The best results are highlighted in \textbf{bold}, second best are \underline{underlined}.}
    \resizebox{1.0\linewidth}{!}{
        \begin{threeparttable}
        \begin{tabular}{ccl|rccccrr|rccccr}
        \toprule
            \rowcolor{lightgray} & & & \multicolumn{7}{c}{\textbf{Human3.6M}~\cite{ionescu2013human3}} & \multicolumn{6}{c} {\textbf{AMASS}~\cite{AMASS2019}} \\

            \multirow{2}{*}{\makecell{Multi \\ Modal}} & \multirow{2}{*}{\makecell{\large UA}} & \multirow{2}{*}{\makecell{{\large Method}}} & Diversity & \multicolumn{2}{c}{Accuracy} & \multicolumn{2}{c}{Multi-modal Accuracy} &\multicolumn{2}{c|}{Plausibility} & Diversity & \multicolumn{2}{c}{Accuracy} & \multicolumn{2}{c}{Multi-modal Accuracy} &\multicolumn{1}{c}{Plausibility} \\
            \cmidrule{4-4} \cmidrule(lr){5-6} \cmidrule(lr){7-8} \cmidrule(lr){9-10}
            \cmidrule(lr){11-11} \cmidrule(lr){12-13} \cmidrule(lr){14-15} \cmidrule(lr){16-16}
        
              & & & APD$\uparrow$ & ADE$\downarrow$ & FDE$\downarrow$ & MMADE$\downarrow$ & MMFDE$\downarrow$ & CMD$\downarrow$ & FID$\downarrow$ &
              APD$\uparrow$ & ADE$\downarrow$ & FDE$\downarrow$ & MMADE$\downarrow$ & MMFDE$\downarrow$ & CMD$\downarrow$ \\ 
        \midrule
             \ding{55} & \ding{55} & TPK~\cite{walker2017pose} & 6.723 & 0.461 & 0.560 & 0.522 & 0.569 & 6.326 & 0.538 & 9.283 & 0.656 & 0.675 & 0.658 & 0.674 & 17.127 \\
             
             \ding{55} & \ding{55} & DLow~\cite{yuan2020dlow} & 11.741 & 0.425 & 0.518 & 0.495 & 0.531 & 4.927 & 1.255 & \underline{13.170} & 0.590 & 0.612 & 0.618 & 0.617 & 15.185 \\
             
             \ding{55} & \ding{55} & GSPS~\cite{mao2021GSPS} & 14.757 & 0.389 & 0.496 & 0.476 & 0.525 & 10.758 & 2.103 & 12.465 & 0.563 & 0.613 & 0.609 & 0.633 & 18.404\\
             
             \ding{55} & \ding{51} & Motron~\cite{salzmann2022motron} & 7.168 & 0.375 & 0.488 & 0.509 & 0.539 & 40.796 & 13.743 & - & - & - & - & - & - \\
             
             \ding{55} & \ding{55} & DivSamp~\cite{dang2022diverse} & 15.310 & 0.370 & 0.485 & 0.477 & 0.516 & 11.692 & 2.083 & \textbf{24.724} & 0.564 & 0.647 & 0.623 & 0.667 & 50.239 \\
             
             \ding{55} & \ding{55} & STARS~\cite{xu2022diverse} & \textbf{15.884} & 0.358 & 0.445 & \underline{0.442} & 0.471 & 15.242 & 1.568 & - & - & - & - & - & - \\
             
             \ding{55} & \ding{55} & MotionDiff \cite{wei2023MotionDiff} & \underline{15.353} & 0.411 & 0.509 & 0.508 & 0.536 & - & - & - & - & - & - & - & - \\
             
             \ding{55} & \ding{55} & HumanMAC~\cite{chen2023humanmac} & 6.301 & 0.369 & 0.480 & 0.509 & 0.545 & - & - & 9.321 & 0.511 & 0.554 & 0.593 & 0.591 & - \\     
             
             \ding{55} & \ding{55} & BeLFusion~\cite{barquero2023belfusion} & 7.602 & 0.372 & 0.474 & 0.473 & 0.507 & 5.988 & 0.209 & 9.376 & 0.513 & 0.560 & 0.569 & 0.585 & 16.995 \\
             
             \ding{55} & \ding{55} & SLD \cite{xu2024learning} & 8.741 & 0.348 & \underline{0.436} & \textbf{0.435} & \textbf{0.463} & 7.418 & 0.930 & - & - & - & - & - & - \\
             
             \ding{55} & \ding{55} & CoMusion~\cite{sun2024comusion} &  7.632 & 0.350 & 0.458 & 0.494 & 0.506 & \underline{3.202} & \underline{0.102} & 10.848 & 0.494 & 0.547 & \textbf{0.469} & \textbf{0.466} & \underline{9.636} \\   
             
              \ding{55} & \ding{55} & TransFusion~\cite{tian2024transfusion} & 5.975 & 0.358 & 0.468 & 0.506 & 0.539 & - & - & 8.853 & 0.508 & 0.568 & 0.589 & 0.606 & - \\
             
             \ding{55} & \ding{51} & ProbHMI~\cite{ma2025probHMI} & 6.682 & 0.364 & 0.493 & 0.511 & 0.558 & - & 0.646 & - & - & - & - & - & - \\    
             
             \ding{55} & \ding{55} & SkeletonDiff~\cite{curreli2025nonisotropic} & 7.249 & \underline{0.344} & 0.450 & 0.487 & 0.512 & 4.178 & 0.123 & 9.456 & \underline{0.480} & \underline{0.545} & 0.561 & 0.580 & 11.417 \\
        \midrule        
             \ding{51} & \ding{51} & Ours & 4.804 & \textbf{0.333} & \textbf{0.399} & 0.471 & \underline{0.464} & \textbf{3.015} & \textbf{0.088} & 7.144 & \textbf{0.461} & \textbf{0.474} & \underline{0.540} & \underline{0.509} & \textbf{8.579} \\
        \bottomrule 
        \end{tabular}
        \end{threeparttable}
    }
    \label{TAB:diverse_quan_result_B50}
\end{table*}

\vspace{-0.4cm}
\paragraph{Evaluation Metrics.}
We evaluate our method from three perspectives: \textit{accuracy}, \textit{diversity} and \textit{plausibility}. For \textit{accuracy}, we employ Best-of-N metrics: 1) \textbf{ADE} (Average Displacement Error), which measures the L2 distance between the most accurate prediction and the ground truth; 2) \textbf{FDE} (Final Displacement Error), similar to ADE but computed only for the final frame; 3) \textbf{MMADE} (Multi-Modal-ADE), a multi-modal extension of ADE in which similar sequences -- grouped based on the L2 distance of the last observed frame -- serve as the multi-modal ground truth; 4) \textbf{MMFDE} (Multi-Modal-FDE), the multi-modal counterpart of FDE. For \textit{diversity}, we use \textbf{APD} (Average Pairwise Distance), which calculates the average distance among all predictions. For \textit{plausibility}, we employ two metrics: 1) \textbf{FID} (Fr\'echet Inception Distance), which evaluates the visual realism of the predicted sequence through features extracted from a pretrained classifier, and 2) \textbf{CMD} (Cumulative Motion Distribution) \cite{barquero2023belfusion}, which quantifies distributional consistency by comparing the area under the cumulative empirical and predicted motion distributions. The detailed definition of these metrics are provided in \appendixref{SEC:metric_detail}.

\subsection{Comparison with the State-of-the-Art}
\subsubsection{Quantitative Results and Analyses}

We compare our approach with stochastic HMP baselines to validate the \textit{accuracy}, \textit{diversity}, and \textit{plausibility} on a set of 50 predicted samples. The baselines include VAE-based methods~\cite{walker2017pose,yuan2020dlow,mao2021GSPS,dang2022diverse,xu2022diverse,xu2024learning}, diffusion-based methods ~\cite{wei2023MotionDiff,barquero2023belfusion,chen2023humanmac,sun2024comusion,tian2024transfusion,curreli2025nonisotropic}, Motron~\cite{salzmann2022motron} and ProbHMI~\cite{ma2025probHMI}. The quantitative results of which are summarized in \cref{TAB:diverse_quan_result_B50}, and our results are computed using an Euler solver with 100 integration steps. Additional results from other ODE solvers and steps, along with a computational efficiency analysis, are provided in \appendixref{SEC:ODE_solver}.

In terms of \textit{accuracy}, our method achieves state-of-the-art performance across most metrics, with notable improvements in ADE and FDE, i.e, 8.5\% and 13\% gains in FDE on the Human3.6M and AMASS datasets, respectively. For the MMADE and MMFDE metrics, our approach consistently ranks among the top performers, securing second or third place on both datasets. For instance, on AMASS, it is only surpassed by CoMusion while maintaining a clear margin over all other baselines.

In terms of \textit{diversity} and \textit{plausibility}, while VAE-based methods (i.e., GSPS~\cite{mao2021GSPS} and STARS~\cite{xu2022diverse}) demonstrates higher APD scores, their predictions often exhibit unrealistic motions, as noted in prior studies~\cite{barquero2023belfusion,chen2023humanmac,curreli2025nonisotropic} and reflected in their substantially worse FID and CMD scores. These findings suggest that such approaches not only lack semantic diversity but also fail to maintain temporal consistency with past observations, thereby hindering their ability to model plausible futures. Similarity, parametric methods~\cite{salzmann2022motron,ma2025probHMI} also struggle with \textit{plausibility} due to their explicit sampling strategies. Although our method shows a decrease in \textit{diversity} compared to diffusion-based models, its competitive performance on MMADE and MMFDE demonstrates its ability to effectively capture the ground-truth distribution. This finding, together with the superior \textit{plausibility} performance of our method, indicates that it generates more meaningful and realistic predictions with efficient coverage~\cite{mohamed2022social,maeda2023fast}, avoiding implausible predictions merely to “guess” the ground truth and artificially increase \textit{diversity}. Since \textit{accuracy} and \textit{plausibility} are all crucial for downstream decision-making, the balanced performance of our method holds significant practical value.

\begin{figure*}[htb]
  \centering
  \resizebox{1.0\linewidth}{!}{      
      \begin{tabular}{c}
          \vspace{0.3cm}
        \begin{tabular}{cc}
            \begin{tabular}{l}
                \subfloat{\includegraphics[width=1.0\linewidth]{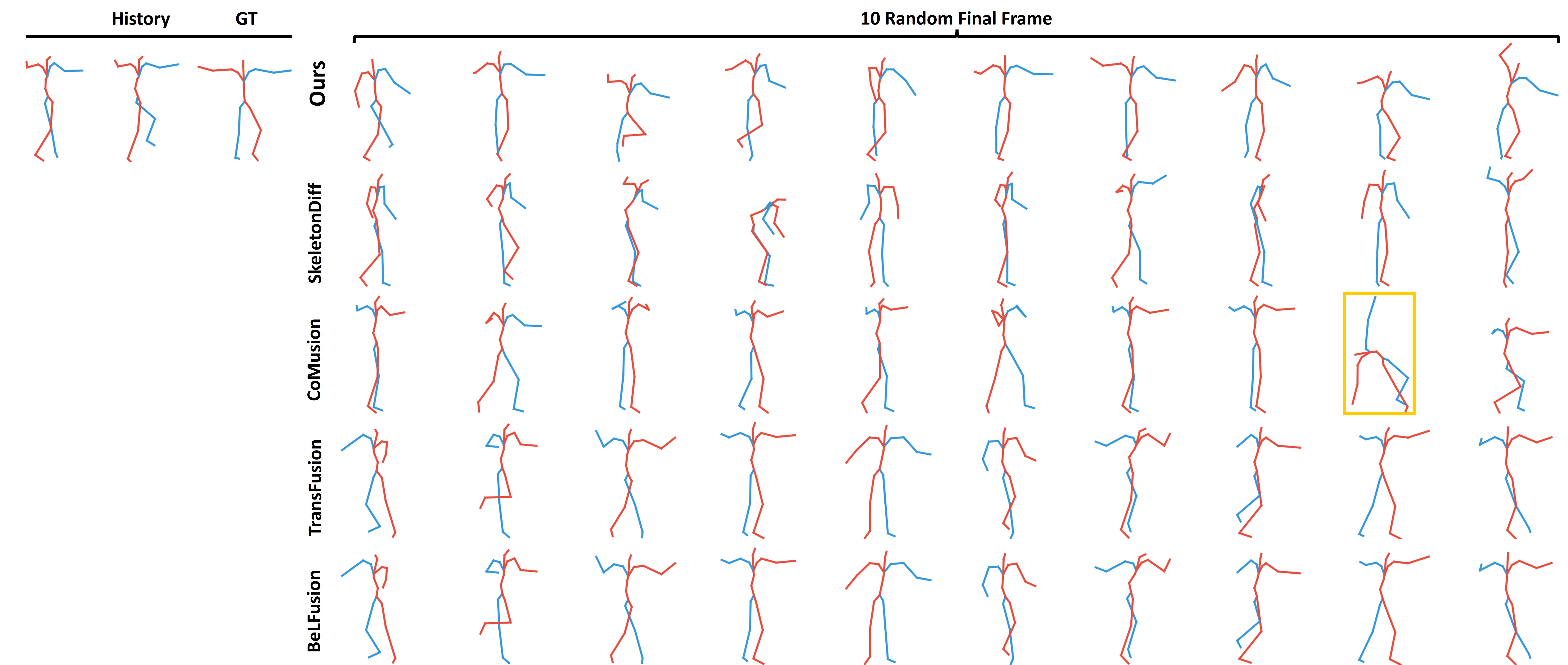}} 
            \end{tabular} &
            
            \begin{tabular}{l}
                \subfloat{\includegraphics[width=1.0\linewidth]{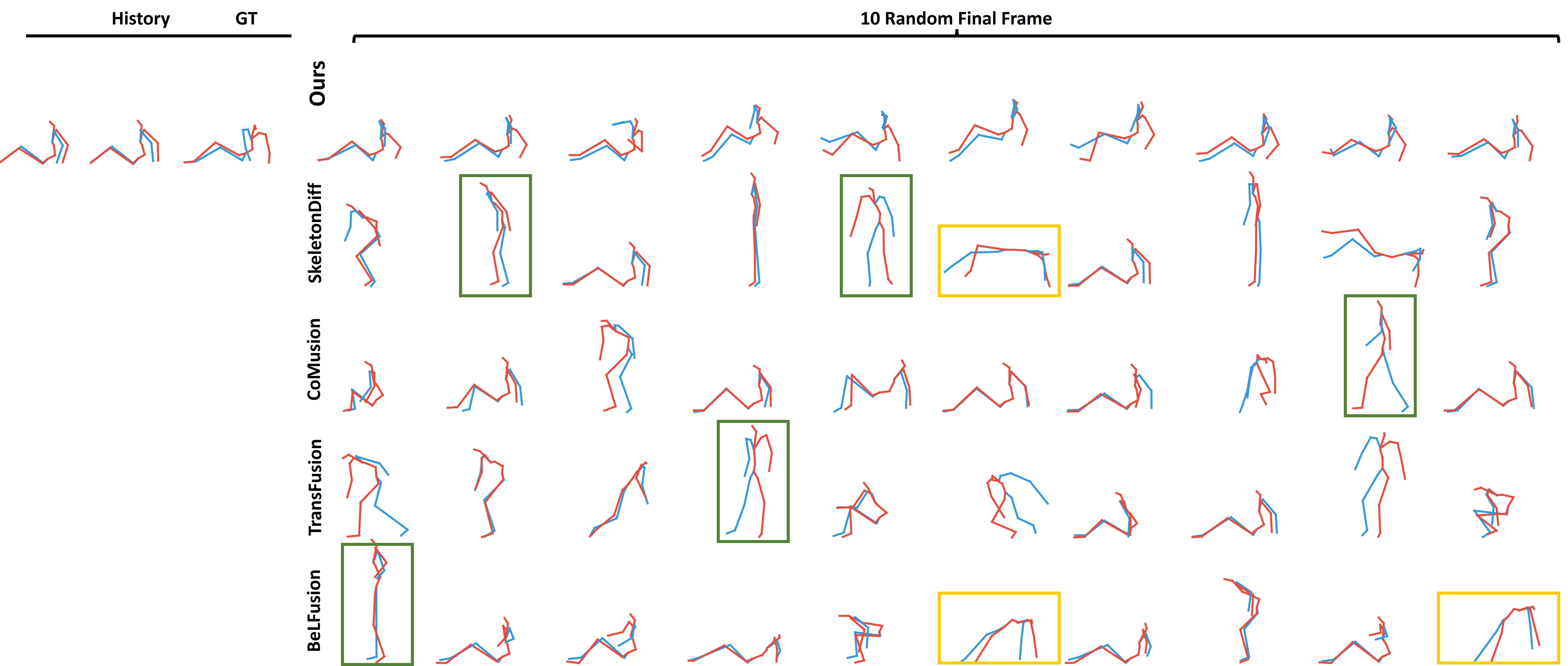}} 
            \end{tabular}
        \end{tabular} \\
        
        \begin{tabular}{cc}
            \begin{tabular}{l}
                \subfloat{\includegraphics[width=1.0\linewidth]{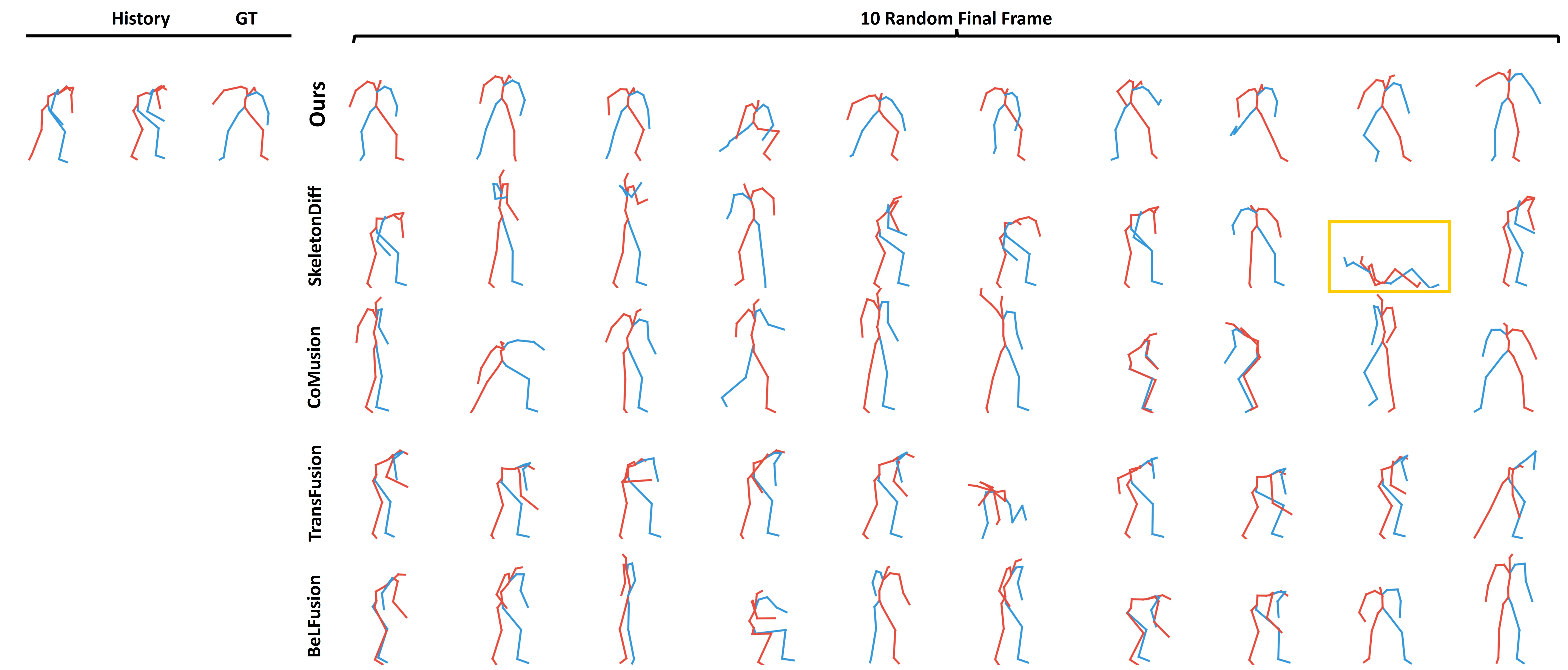}} 
            \end{tabular} &
            
            \begin{tabular}{l}
                \subfloat{\includegraphics[width=1.0\linewidth]{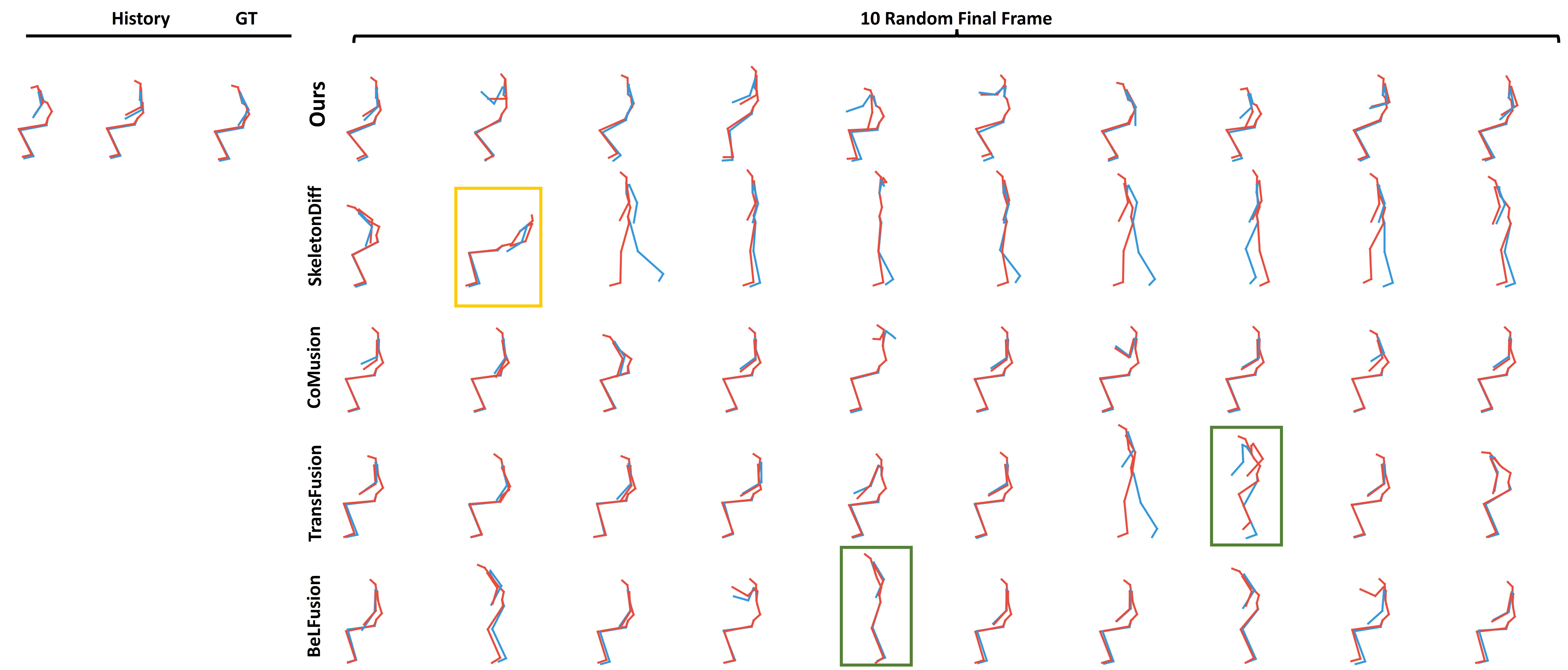}} 
            \end{tabular}
        \end{tabular} 
      \end{tabular}
  }
\caption{Qualitative results. We present qualitative comparison results with SkeletonDiff~\cite{curreli2025nonisotropic}, CoMusion~\cite{sun2024comusion}, TransFusion~\cite{tian2024transfusion}, and BeLFusion~\cite{barquero2023belfusion} on the AMASS dataset. For each method, we visualize the final frame of 10 randomly sampled predictions.} 
\label{fig:qualitative_results}
\end{figure*}

\subsubsection{\my{Qualitative Results and Analyses}}
We present qualitative comparisons against several recent SOTA methods, including BeLFusion~\cite{barquero2023belfusion}, TransFusion~\cite{tian2024transfusion}, CoMusion~\cite{sun2024comusion}, and SkeletonDiff~\cite{curreli2025nonisotropic}, on the AMASS dataset, all of which 
demonstrate strong performance in terms of \textit{accuracy}, \textit{diversity} and \textit{plausibility}. Although these approaches generate diverse motion predictions, our method produces more natural and realistic pose forecasts, exhibiting higher plausibility and better alignment with the semantics of the observed history. For example, in the top-right group, diffusion-based baselines occasionally generate extreme joint bends (highlighted in yellow), whereas such artifacts are absent in our results. They also sometimes violate historical motion semantics (highlighted in green). Similar artifacts can be observed in other visualizations. While our method yields slightly lower \textit{diversity} than the baselines, the 10 randomly sampled predictions still demonstrate meaningful variation. For instance, in the dance sequence shown in the top-left group, our method produces noticeable differences in arm swing amplitude and timing, indicating that it preserves non-trivial diversity when the motion history permits it. Overall, these results highlight the trade-off between \textit{diversity} and \textit{plausibility} in our method: it maintains diversity for flexible and complex actions such as dancing, while avoiding excessive speculation for more constrained actions such as sitting, thereby prioritizing plausibility. Additional examples are provided in \appendixref{SEC:additional_visualization} and the supplementary video. 

\subsection{Ablation Studies}
We begin with the ablation study on the construction of the latent space, as summarized in \cref{TAB:ablation_study}. First, we replace the latent distribution $q_{z}(\mathbf{Z})$ with a mixture model composed of fixed sub-Gaussians. Specifically, the fixed assignment centers are initialized through latent space clustering via model parameters obtained after their initial configuration, following the strategy adopted in MGF~\cite{chen2024mgf}. Technically, this is equivalent to train the normalizing flow model using \cref{EQ:flow_model_update} in isolation. Second, we adopt a single-modal Gaussian (i.e., a standard normal distribution) as the target latent distribution, which is commonly used in prior latent model-based approaches~\cite{barquero2023belfusion,mao2021GSPS,ma2025probHMI}. Finally, we remove the latent space entirely and utilize the neural ODE for motion prediction directly in the original data space.

As illustrated in the middle part of \cref{TAB:ablation_study}, when $q_{z}(\mathbf{Z})$ is defined as a fixed mixture distribution (w/ Fixed Gaussian), most metrics exhibit performance degradation, except for CMD on Human3.6M and APD on AMASS. These results suggest that a learnable distribution can capture the underlying data distribution more accurately, owing to its ability to adapt throughout training. When $q_{z}(\mathbf{Z})$ is defined as a standard Gaussian (w/ Standard Normal) or entirely removed (w/ FM only), the performance deteriorates even further, particularly on  metrics on \textit{plausibility}. This demonstrate that a multi-modal prior is more effective for HMP than a conventional single-modal prior, as it can better capture the diverse patterns and disentangle semantic representations by modeling multiple sub-distributions, thereby enabling more plausible and accurate motion forecasting. We further provide a visualization of the latent space using t-SNE in \appendixref{SEC:vis_latent_space} to support our findings.

Next, we conduct ablation studies to investigate the design choices of the latent dynamics model. First, we replace the source distribution, $p_{\mathbf{Z}_{0}}$, from $\mathcal{N}(\mathbf{Z}_{0}, \mathrm{I})$ to a standard Gaussian distribution $\mathcal{N}(0, \mathrm{I})$. Second, we remove the observation $\mathbf{x}_{obs}$ from the model, which serve as the condition, resulting in a reconstruction architecture. Further, we substitute our temporal tokenization scheme with a conventional spatial tokenization strategy, following prior latent diffusion-based methods~\cite{chen2023humanmac,sun2024comusion}, where each transformer token corresponds to the motion at a single time step or a single temporal frequency.

As illustrated in the bottom part of \cref{TAB:ablation_study}, when employing $p_{\mathbf{Z}_{0}}$ as $\mathcal{N}(0, \mathrm{I})$ (w/o Past), the \textit{diversity} increases slightly on both datasets but exhibits a significant degradation in \textit{accuracy}. Regarding \textit{plausibility}, the CMD score on Human3.6M improves marginally, whereas the CMD score on AMASS and the FID score on Human3.6M deteriorate, indicating inferior \textit{plausibility} compared to $\mathcal{N}(\mathbf{Z}_{0}, \mathrm{I})$. When the condition $\mathbf{x}_{obs}$ is removed (w/o Condition), the model exhibits decreases in both \textit{diversity} and \textit{accuracy}. In terms of \textit{plausibility}, only the CMD score on Human3.6M shows a marginal improvement, while all other metrics decrease. This finding demonstrates that explicitly conditioning on the observed history provides more effective guidance than treating it solely as a reconstruction target. When using spatial tokenization (w/ Spatial Tokenization), both \textit{accuracy} and \textit{plausibility} metrics degrade substantially compared to the fair-comparison variant without explicit conditioning (w/o Condition). This result validates the effectiveness of our skeleton-aware transformer model, which incorporates structural priors for human motion modeling.

\begin{table*}[htb]
    \renewcommand{\arraystretch}{1.25}
    \centering
    \caption{The results of ablation studies on the Human3.6M and AMASS datasets. Since AMASS does not include action labels, FID is not used for evaluation. The best results are highlighted in \textbf{bold}.}
    \resizebox{1.0\linewidth}{!}{
        \begin{threeparttable}
        \begin{tabular}{l|rccccrr|rccccr}
        \toprule
            & \multicolumn{7}{c|}{\textbf{Human3.6M}~\cite{ionescu2013human3}} & \multicolumn{6}{c} {\textbf{AMASS}~\cite{AMASS2019}} \\        
              & APD$\uparrow$ & ADE$\downarrow$ & FDE$\downarrow$ & MMADE$\downarrow$ & MMFDE$\downarrow$ & CMD$\downarrow$ & FID$\downarrow$ &
              APD$\uparrow$ & ADE$\downarrow$ & FDE$\downarrow$ & MMADE$\downarrow$ & MMFDE$\downarrow$ & CMD$\downarrow$ \\ 
        \midrule    
             Ours & 4.804 & \textbf{0.333} & \textbf{0.399} & \textbf{0.471} & \textbf{0.464} & 3.015 & \textbf{0.088} & 7.144 & \textbf{0.461} & \textbf{0.474} & \textbf{0.540} & \textbf{0.509} & \textbf{8.579} \\
        \midrule
             w/ Fixed Gaussian & 4.676 & 0.336 & 0.407 & 0.478 & 0.475 & \textbf{2.767} & 0.133 & 7.311 & 0.463 & 0.474 & 0.541 & 0.509 & 9.583 \\
             w/ Standard Normal & 5.056 & 0.356 & 0.421 & 0.483 & 0.479 & 5.854 & 0.143 & 7.339 & 0.465 & 0.477 & 0.545 & 0.511 & 10.706 \\
             w/ FM Only & 4.553 & 0.341 & 0.408 & 0.480 & 0.475 & 4.895 & 0.156 &  7.521 & 0.469 & 0.475 & 0.546 & 0.509 & 12.563 \\
        \midrule
            w/o Past & 5.008 & 0.346 & 0.419 & 0.478 & 0.478 & 2.917 & 0.223 & \textbf{7.531} & 0.479 & 0.486 & 0.557 & 0.513 & 8.849 \\
            w/o Condition & 4.791 & 0.352 & 0.417 & 0.482 & 0.479 & 2.851 & 0.118 & 7.030 & 0.485 & 0.482 & 0.544 & 0.510 & 8.981 \\
            w/ Spatial Tokenization & \textbf{5.691} & 0.397 & 0.461 & 0.509 & 0.517 & 3.896 & 0.208 & 7.505 & 0.498 & 0.502 & 0.553 & 0.527 & 10.056 \\
        \bottomrule 
        \end{tabular}
        \end{threeparttable}
    }
    \label{TAB:ablation_study}
\end{table*}


\subsection{\my{Analysis of Likelihood Estimation}} \label{SEC:likelihood_estimation}
To validate the effectiveness of our method, particularly the proposed Gaussian-mixture prior, in likelihood estimation, we utilize the \textbf{log-likelihood} metric and compare our method with three variants: 1) our method using a standard normal prior (w/ Standard Normal), 2) our method without any prior (w/ FM Only), and 3) the baseline method Motron~\cite{salzmann2022motron}, which adopts a parametric representation based on a concentrated Gaussian in $\mathrm{SO}(3)$. Given a ground truth sequence $\mathbf{X}$, the \textbf{log-likelihood} metric of $\mathbf{X}$ is defined as:
\begin{equation}
    \mathrm{LL} = \log p_{\theta}(\mathbf{X}),
\end{equation}
\noindent where $p_{\theta}$ denotes the predicted probability distribution parameterized by each method under comparison. 

\begin{table}[h]
    \centering
    \renewcommand{\arraystretch}{1.25}
    \caption{The results of \textbf{log-likelihood}$\uparrow$ on the Human3.6M and AMASS datasets.}
    \resizebox{1.0\linewidth}{!}{
        \begin{tabular}{c|c|c|c|c}
        \toprule
         \diagbox{Dataset}{Method} & Ours & w/ FM Only & w/ Standard Normal & Motron~\cite{salzmann2022motron} \\
        \midrule
        Human3.6M & \textbf{1.584} & 0.777 & -0.134 & 1.319 \\ 
        \hline
        AMASS & \textbf{-0.628} & -0.677 & -1.105 & - \\ 
        \bottomrule
        \end{tabular}
    }
    \label{TAB:likelihood_estimatoin}
\end{table}

As illustrated in \cref{TAB:likelihood_estimatoin}, our method with the Gaussian-mixture prior achieves the best LL performance on both datasets, demonstrating its effectiveness in modeling complex data distributions. Notably, the variant without any prior (w/ FM Only) even outperforms the version using a standard normal prior (w/ Standard Normal), indicating that imposing an overly simplistic prior on complex human motion distributions can negatively affect distribution modeling. Furthermore, our method also surpasses Motron, highlighting the advantages of our probabilistic formulation. This strong probabilistic modeling capability further supports the reliability of using likelihood as an uncertainty quantification measure in our method. Additional quantitative and qualitative evaluations for uncertainty quantification are presented in \appendixref{SEC:quantitative_evaluation_uncertainty} and \appendixref{SEC:additional_visualization}.

\section{Conclusion}
We present a latent flow-based model for stochastic HMP, equipped with a data-driven Gaussian mixture prior that adaptively captures diverse motion patterns in the latent space. Our method achieves state-of-the-art performance on two large-scale datasets. Additionally, our model is fully invertible, enabling tractable likelihood estimation for natural uncertainty quantification.

\vspace{-0.3cm}
\paragraph{Limitations and Future Work}
Our method still has several limitations. First, although the constructed Gaussian mixture prior is effective for motion prediction, its direct application to other tasks, such as classification~\cite{izmailov2020semi} and anomaly detection~\cite{hirschorn2023normalizing}, yields suboptimal performance. This limitation partly arises from the lack of labeled data to supervise the clustering process, and partly from the inherent difficulty of constructing a latent space with both clear semantic boundary for classification and sufficient continuity for generation. Second, although our prior construction approach can adaptively adjust to the dataset, it may perform poorly when encountering long-tailed data distributions. In such cases, the latent mixture distribution may degenerate into a single-modal form to fit the dominant data mode. While several techniques, such as resampling-based methods, have been proposed to address this problem, they are difficult to apply in our framework due to the absence of labels. Imposing additional constraints on the assignment process may offer a potential solution, which we plan to explore in future work. \my{Finally, although we believe that, in the absence of additional guidance, preserving the known motion context is generally preferable, we acknowledge that strictly adhering to the semantics inferred from the observed history can be conservative. In reality, humans may change their motion patterns in the near future, and such deviations may not be captured by a history-consistent prediction.}

\section*{Acknowledgments}
\vspace{-0.1cm}
This work was supported by the National Natural Science Foundation of China (Ref: 62272019, Liang).

{
    \clearpage
    \small
    \bibliographystyle{ieeenat_fullname}
    \bibliography{main}
}

\appendix
\clearpage
\setcounter{page}{1}
\setcounter{section}{0}
\setcounter{equation}{0}

\renewcommand\thesection{\Roman{section}}
\renewcommand\theequation{\thesection.\arabic{equation}}

\maketitlesupplementary

\section{Metric Calculation Details} \label{SEC:metric_detail}
Followings \cref{SEC:problem_formulation}, the observed $T$ frames are represented as $\mathbf{X}_{obs}$ = $(\mathbf{x}_{1}, \cdots, \mathbf{x}_{T})$, and the future motion with $N$ steps is denoted as $\mathbf{Y}$ = $(\mathbf{x}_{T+1}, \cdots, \mathbf{x}_{T+N})$. The diverse set of $M$ predicted future samples is represented as $\hat{\mathbf{Y}}^M$= $\{\hat{\mathbf{Y}}^{1}, \cdots, \hat{\mathbf{Y}}^{M}\}$. The metrics used in the experiments can then be defined as follows:

\begin{itemize}
    \item \textbf{Average Pair Distance (APD)} measures the L2 distance between a set of predictions generated from the same history, which is computed as:
    \begin{equation}
            \mathrm{APD} = \frac{1}{M(M-1)} \sum_{i=1}^M \sum_{j \neq i}^{M} \| \hat{\mathbf{Y}}^i - \hat{\mathbf{Y}}^j \|_2.
    \end{equation}    
    
    \item \textbf{Average Displacement Error (ADE)} computes the average L2 distance between the ground truth and the closest sample, which is computed as:
    \begin{equation}
            \mathrm{ADE} = min_{\hat{\mathbf{Y}}^j} \| \mathbf{Y} - \hat{\mathbf{Y}}^j \|_2.
    \end{equation}
    
    \item \textbf{Final Displacement Error (FDE)} computes the L2 distance between the final pose of the GT and the closest final pose among predictions, which is computed as:
    \begin{equation}
            \mathrm{FDE} =  min_{\hat{\mathbf{x}}_{T+N}^j}\| \mathbf{x}_{T+N} - \hat{\mathbf{x}}_{T+N}^j \|_2.
    \end{equation}
    
    \item \textbf{Multi-Modal-ADE (MMADE)} aggregates multi-modal ground truths (MMGT) across the entire dataset. Given a dataset $D$ containing $M$ sequences, the MMGT of $\mathbf{X}_{obs}^i \in D$ is then computed as \cref{EQ:mmgt}: 
    \begin{equation}
        \mathrm{MMGT} = \left\{ \mathbf{Y}^j \mid \| \mathbf{x}_T^j - \mathbf{x}_T^i \|_2 < \mathrm{A}, \,  j = 1, 2, \dots, M \right\},
    \label{EQ:mmgt}
    \end{equation}
    where $\mathrm{A}$ is a hyperparameter, often set as 0.5 for both Human3.6M and HumanEva-\uppercase\expandafter{\romannumeral1} datasets. MMADE is then computed as \cref{EQ:MMADE}:
    \begin{equation}
            \mathrm{MMADE} = \frac{1}{L} \sum_{j=1}^{L} min_{\hat{\mathbf{Y}}^k} \| \mathbf{Y}^{j} - \hat{\mathbf{Y}}^k \|_2,
    \label{EQ:MMADE}
    \end{equation}
    where $\mathrm{x}_{j,\cdot}$ represents $j$-th sequence belonging to the corresponding MMGT and $L$ denotes the number of sequences contained in the MMGT.

    \item \textbf{Multi-Modal-FDE (MMFDE)} is computed using the same MMGT as MMADE and is defined as:
    \begin{equation}
            \mathrm{MMFDE} = \frac{1}{L} \sum_{j=1}^{L} min_{\hat{\mathbf{x}}_{T+N}^k} \| \mathbf{x}_{j,T+N} - \hat{\mathbf{x}}_{T+N}^k \|_2.
    \label{EQ:MMFDE}
    \end{equation}

    \item \textbf{Cumulative Motion Distribution (CMD)} measures the difference between the areas under the cumulative pseudo-data motion distribution and the predicted distribution. Let $\bar{D}$ denote the L2 distance between 
    the displacement in two consecutive frames across the entire dataset. For the $f$-th frame in the predicted distribution, we compute the average displacement $D_j$ in the same manner. The CMD is then calculated as:
    \begin{equation}
        \begin{split}
            \mathrm{CMD} &= \sum_{i=1}^{N-1} \sum_{j=1}^{i} \| D_j - \bar{D} \|_{1} \\
            & = \sum_{i=1}^{N-1} (N-i) \| D_i - \bar{D} \|_{1}.
        \end{split}
    \label{EQ:CMD}
    \end{equation}
\end{itemize}

\begin{figure}[htbp]
    \centering
    \begin{tabular}{cc}
    \vspace{0.2cm}
    \subfloat[Colored according to the corresponding sub-distribution]{
        \includegraphics[width=0.85\linewidth]{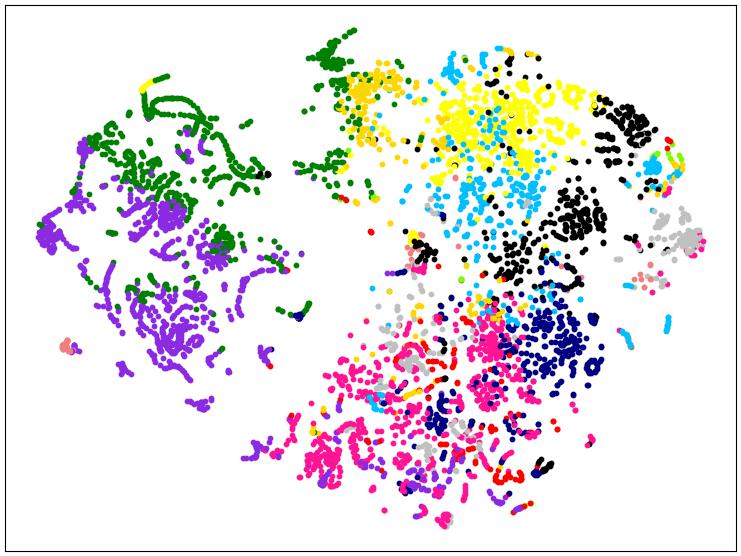}    
        \label{fig:plot_sub_distribution}
    } \\
    \subfloat[Colored according to the corresponding class label]{
        \includegraphics[width=0.85\linewidth]{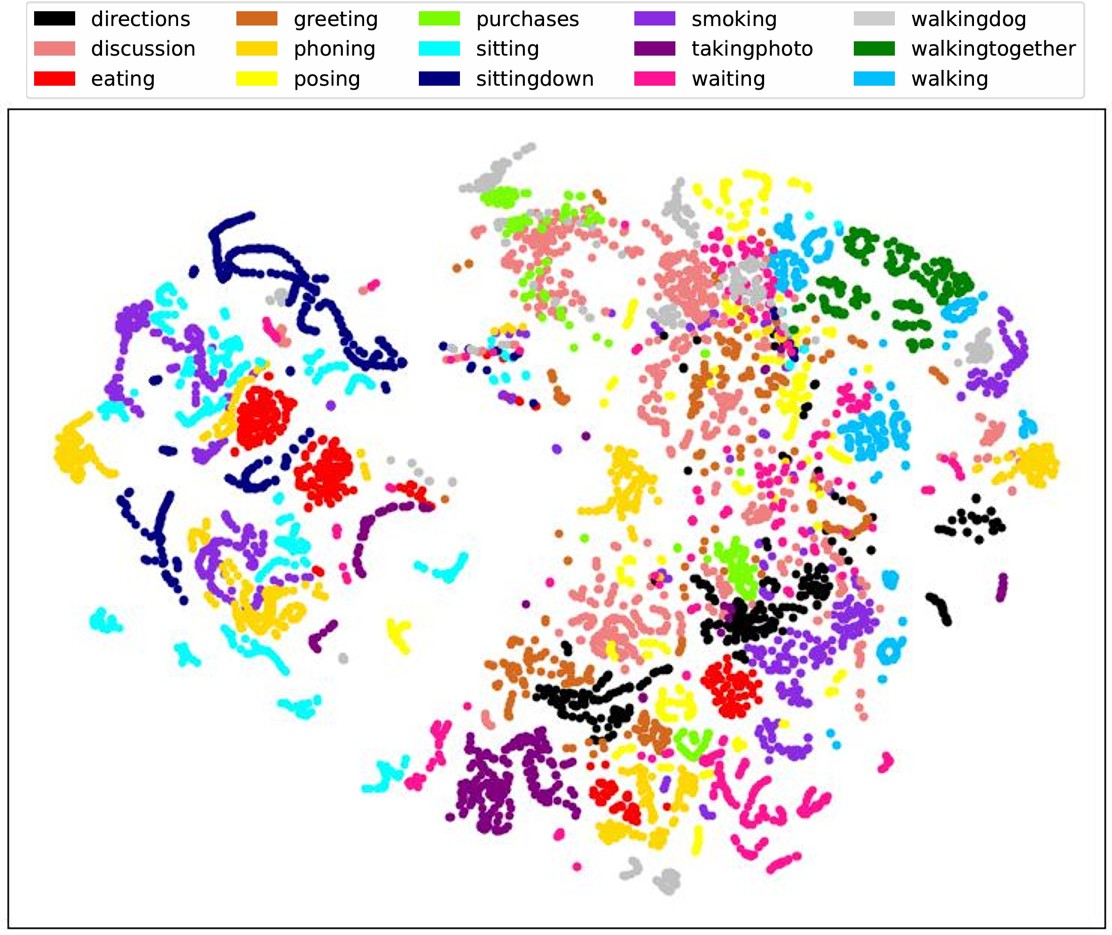}    
        \label{fig:plot_class_label}
    }
    \end{tabular}
    \caption{Visualization of the multi-modal latent space. We present a 2D t-SNE projection of the latent encodings for all Human3.6M test sequences: (a) colored by their nearest sub-distribution, and (b) colored by the action label of each sequence.}
    \label{fig:latent_visualization_multi_modal}
\end{figure}

\begin{figure}[htbp]
    \centering
    \begin{tabular}{cc}
    \subfloat[Colored according to the corresponding class label]{
        \includegraphics[width=0.85\linewidth]{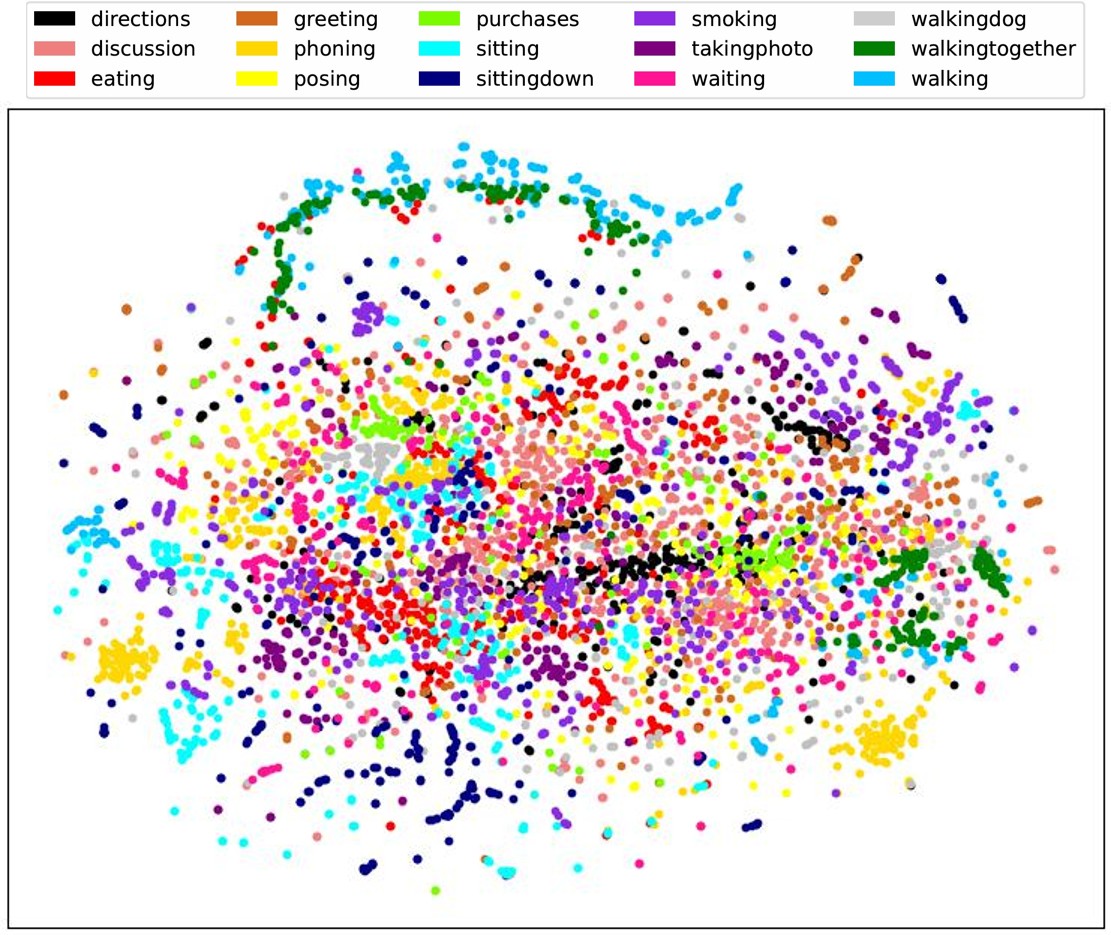}    
        \label{fig:plot_class_label_single_modal}
    }
    \end{tabular}
    \caption{Visualization of the single-modal latent space. We present a 2D t-SNE projection of the latent encodings for all Human3.6M test sequences, which is colored by the action label.}
    \label{fig:latent_visualization_single_modal}
\end{figure}

\section{Visualization of the Latent Space} \label{SEC:vis_latent_space}
We present a t-SNE visualization of the multi-modal latent space across all Human3.6M test sequences, as shown in \cref{fig:latent_visualization_multi_modal}. In \cref{fig:plot_sub_distribution}, points sharing the same color correspond to the same sub-distribution, while in \cref{fig:plot_class_label}, points with the same color indicate sequences belonging to the same action class.

From \cref{fig:plot_sub_distribution}, we can observe that the latent space is organized into several clusters. Interestingly, the number of clusters is smaller than the initial setting of 16 due to the automatic pruning mechanism in the employed updating strategy, which performs a soft allocation of samples. Furthermore, these clusters exhibit partial overlap without distinct boundaries between them. This continuity in the latent space is advantageous for generative tasks, as it facilitates smooth transitions between modes, but it can hinder performance in discriminative tasks such as classification and anomaly detection. From \cref{fig:plot_class_label}, we can observe that sequences belonging to the same action class tend to cluster together. Not all sequences of the same class are grouped in a single cluster, primarily because the Human3.6M dataset provides only coarse action labels, namely individual sequences may contain motions from other classes. For example, \textit{Waiting} sequences often include sub-sequences in which the person walks or sits down. Nevertheless, we can also observe that action classes with similar motion semantics, such as \textit{Walking} and \textit{WalkingTogether}, are positioned closely in the latent space, both appearing in the upper-right region. This finding indicates that our constructed multi-modal latent prior effectively captures diverse motion patterns and their underlying semantic relationships.

For comparison, we also provide a t-SNE visualization of the single-modal latent space for all Human3.6M test sequences in \cref{fig:latent_visualization_single_modal}. When compared with \cref{fig:plot_class_label} and \cref{fig:plot_class_label_single_modal}, we observe that sequences of different action labels in the single-modal latent space are largely entangled. In contrast, sequences belonging to the same class form more coherent clusters in our multi-modal latent space, indicating that the proposed multi-modal prior effectively disentangles motion patterns and semantics. Interestingly, a subset of \textit{Walking} and \textit{WalkingTogether} sequences forms clusters outside the main region, a phenomenon also reported in BeLFusion~\cite{barquero2023belfusion}.

\section{Analysis of the Latent Space Construction}
We further examine the influence of the number of initial components in the latent mixture distribution. As shown in \cref{TAB:init_num_latent}, we evaluate configurations with 8, 16, and 32 initial components on Human3.6M. For comparison, we additionally report a variant of the multi-modal prior that employs a single learnable component, as well as the results obtained using a single-modal prior.

\begin{table}[htb]
    \renewcommand{\arraystretch}{1.25}
    \centering
    \caption{Analysis of varying the number of initialized sub-distributions used to construct the latent space, with results reported on Human3.6M.}
    \resizebox{1.0\linewidth}{!}{
        \begin{threeparttable}
        \begin{tabular}{c|c|ccccccc}
        \toprule
            Prior & Nums & APD$\uparrow$ & ADE$\downarrow$ & FDE$\downarrow$ & MMADE$\downarrow$ & MMFDE$\downarrow$ & CMD$\downarrow$ & FID$\downarrow$ \\
        \midrule
        Single & - & 5.056 & 0.356 & 0.421 & 0.483 & 0.479 & 5.854 & 0.143 \\
        \midrule
        MM & 1 & 4.691 & 0.338 & 0.408 & 0.476 & 0.475 & 3.215 & 0.154 \\
        MM & 8 & 4.773 & 0.334 & 0.403 & 0.473 & 0.469 & 3.038 & 0.134 \\
        MM & 16 & 4.804 & 0.333 & 0.399 & 0.471 & 0.464 & 3.015 & 0.088 \\
        MM & 32 & 4.885 & 0.333 & 0.400 & 0.475 & 0.470 & 2.955 & 0.092 \\
        \bottomrule
        \end{tabular}
        \end{threeparttable}
    }
    \label{TAB:init_num_latent}
\end{table}

When the number of initial components is set to 16 or 32, the predictive performance across \textit{accuracy}, \textit{diversity}, and \textit{plausibility} remains nearly identical. However, when the initial number is reduced to 8 or 1, the \textit{accuracy} and \textit{diversity} metrics remain comparable, but the \textit{plausibility} metric deteriorates significantly. These observations indicate that our data-driven Gaussian mixture can effectively and adaptively capture the diverse motion patterns in the dataset once the initial number of components is sufficiently large. In such cases, the soft-alignment strategy prunes redundant components during training, making additional components unnecessary. In contrast, when the initial number of components is too small, the limited capacity restricts the model’s ability to fully capture the inherent diversity of motion patterns. Interestingly, when directly comparing the learnable single-modal prior (with one initial component) to the pre-defined standard Normal prior, the results show that the learnable prior achieves substantially more accurate predictions and yields a much lower CMD value, indicating a closer fit to the underlying data distribution and greater distributional consistency.

\section{Analysis of the ODE Solver} \label{SEC:ODE_solver}
We report predictive performance and computational cost using an Euler solver with integration steps ranging from 1 to 200, as well as a Dopri5 solver with adaptive step sizes. As shown in \cref{TAB:solver_steps}, the Euler solver exhibits convergence when the step count becomes sufficiently large. Specifically, beyond 100 steps, the \textit{accuracy} metrics remain stable, while APD improves only marginally. Moreover, the results closely match those obtained using Dopri5, further confirming convergence behavior. Interestingly, for the Euler solver, when the number of steps is insufficient, some metrics appear slightly better, likely due to stochastic variations introduced during the numerical integration process.

We also report the inference time for each configuration in \cref{TAB:solver_steps}, where every measurement is averaged over 1,000 runs on one RTX 4060Ti GPU to ensure fair comparison. Notably, the Dopri5 solver with adaptive step sizes achieves comparable prediction quality at significantly lower computational cost compared to the Euler solver.


\begin{table}[htb]
    \renewcommand{\arraystretch}{1.25}
    \centering
    \caption{ Experimental results on various ODE solvers.}
    \resizebox{1.0\linewidth}{!}{
        \begin{threeparttable}
        \begin{tabular}{c|c|r|rcccc}
        \toprule
            \multirow{2}{*}{Method} & \multirow{2}{*}{Steps} & \multirow{2}{*}{Time} & \multicolumn{5}{c}{\textbf{Human3.6M}~\cite{ionescu2013human3}} \\
            & & & APD$\uparrow$ & ADE$\downarrow$ & FDE$\downarrow$ & MMADE$\downarrow$ & MMFDE$\downarrow$ \\
        \midrule
        Dopri5 & - & 504ms & 5.046 & 0.335 & 0.401 & 0.472 & 0.465 \\
        \midrule
        Euler & 1 & 16ms & 1.196 & 0.367 & 0.513 & 0.513 & 0.571 \\
        Euler & 10 & 88ms & 3.427 & 0.330 & 0.409 & 0.476 & 0.477 \\
        Euler & 20 & 174ms & 4.069 & 0.330 & 0.400 & 0.472 & 0.468 \\
        Euler & 100 & 886ms & 4.804 & 0.333 & 0.399 & 0.471 & 0.464 \\
        Euler & 200 & 1676ms & 4.915 & 0.334 & 0.400 & 0.471 & 0.464 \\

        \bottomrule
        \toprule
            \multirow{2}{*}{Method} & \multirow{2}{*}{Steps} & \multirow{2}{*}{Time} & \multicolumn{5}{c}{\textbf{AMASS}~\cite{AMASS2019}} \\
            & & & APD$\uparrow$ & ADE$\downarrow$ & FDE$\downarrow$ & MMADE$\downarrow$ & MMFDE$\downarrow$ \\
        \midrule
        Dopri5 & - & 965ms & 7.370 & 0.463 & 0.475 & 0.542 & 0.510 \\ 
        \midrule
        Euler & 1 & 18ms & 0.989 & 0.485 & 0.621 & 0.568 & 0.648 \\
        Euler & 10 & 115ms & 4.123 & 0.446 & 0.489 & 0.531 & 0.525 \\
        Euler & 20 & 223ms & 6.363 & 0.453 & 0.472 & 0.534 & 0.508 \\
        Euler & 100 & 1132ms & 7.144 & 0.461 & 0.474 & 0.540 & 0.509 \\
        Euler & 200 & 2246ms & 7.257 & 0.462 & 0.473 & 0.541 & 0.509 \\
        \bottomrule 
        \end{tabular}
        \end{threeparttable}
    }
    \label{TAB:solver_steps}
\end{table}

\section{Analysis of Uncertainty Quantification} \label{SEC:quantitative_evaluation_uncertainty}

\begin{figure*}[tbp]
    \centering
    \resizebox{0.9\linewidth}{!}{
        \begin{tabular}{ll}
        \subfloat[Probability Rank on Human3.6M]{
            \includegraphics[width=0.45\linewidth]{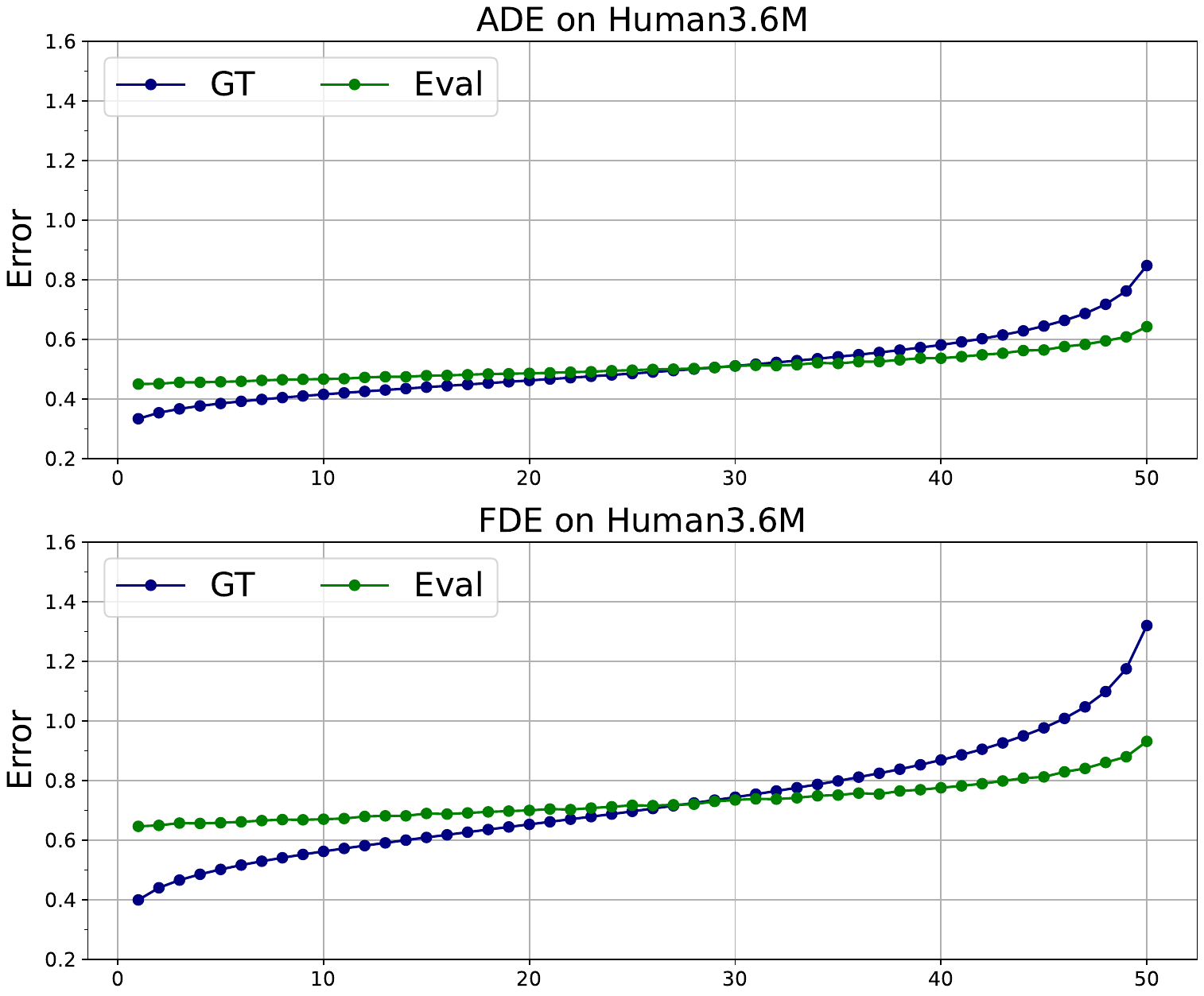}    
            \label{fig:uncertainty_h36m}
        } &
        \subfloat[Probability Rank on AMASS]{
            \includegraphics[width=0.45\linewidth]{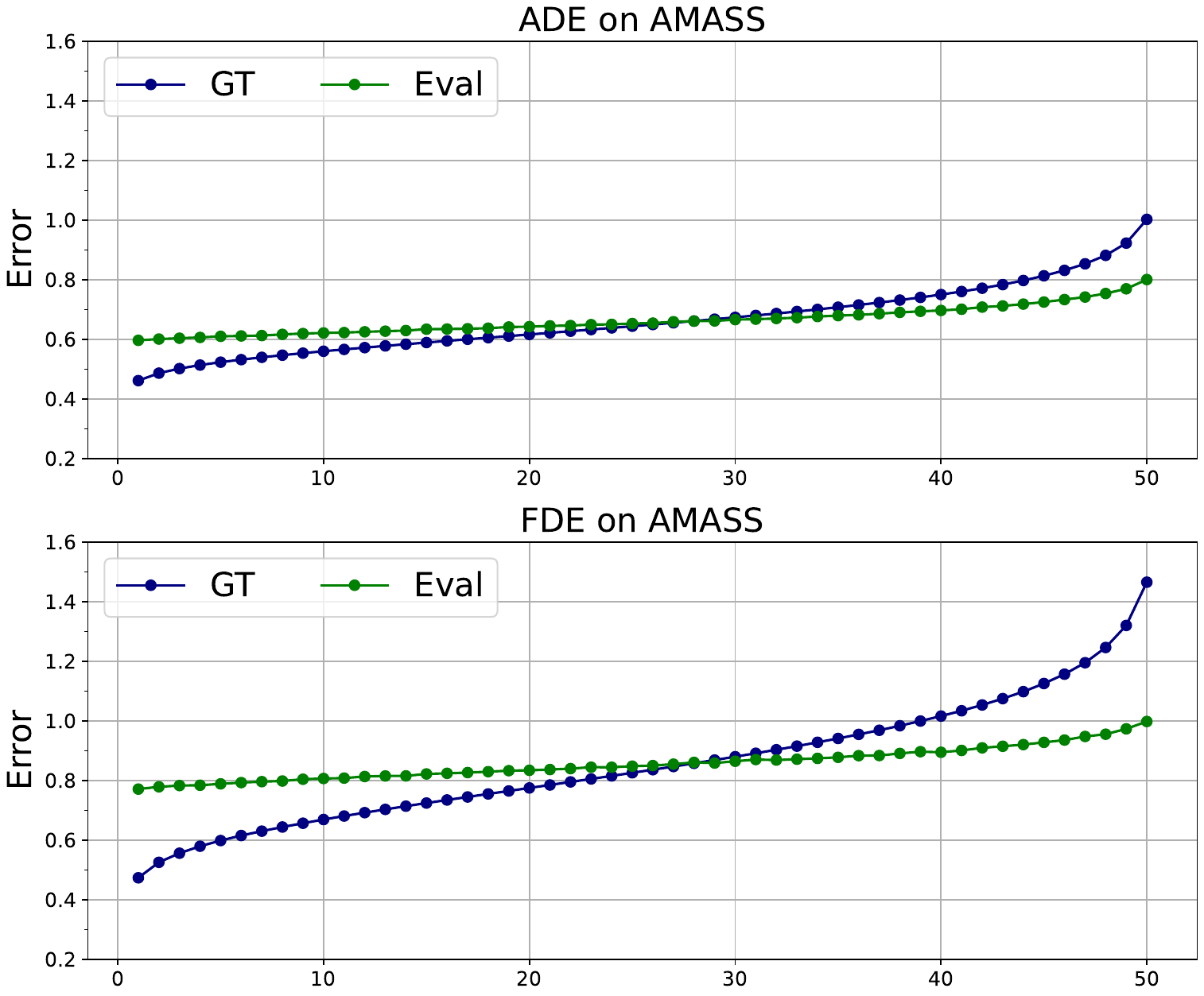}    
            \label{fig:uncertainty_rank_amass}
        }
        \end{tabular}
    }
    \caption{We evaluate the correlation between prediction errors (\textbf{ADE} and \textbf{FDE}) and their associated uncertainty ranks (denoted "Eval") on Human3.6M and AMASS. As a reference benchmark, we also present the values based on ideal ranks (denoted "GT").}
    \label{fig:uncertainty_rank}
\end{figure*}

To validate the reliability of the estimated probabilities as measures of uncertainty, we first examine the correlation between prediction errors (e.g., ADE and FDE) and their corresponding uncertainty quantification ranks, where the ranking is determined by the estimated probability -- samples with higher probability are assigned higher ranks. For comparison, we also include results based on ideal ranks, where the ranking is determined directly by the prediction errors -- samples with lower errors are assigned higher ranks. Ideally, these two rankings should coincide.

As shown in \cref{fig:uncertainty_rank}, while the estimated ranks are not perfectly aligned with the ground truth, they reveal a clear trend that higher probability values (i.e., higher ranks) correlate strongly with lower ADE and FDE errors. This correlation is consistent with statistical theory, confirming that more probable predictions are, on average, more accurate. These findings validate the reliability and practical value of our uncertainty quantification, as predictions assigned higher likelihood by our model are indeed closer to the true future. This enables autonomous systems to effectively prioritize actions based on quantified confidence. 

To quantify the deviation between our estimated ranks and the ideal ranks, we measure the normalized area difference between the estimated error-rank curve and the ideal curve. This metric quantifies the overall discrepancy in the error distribution across all confidence ranks, analogous to the Area Under the Sparsification Error (AUSE)~\cite{lind2024uncertainty}. The formulation of our metric is defined as:
\begin{equation}
E = \frac{\sum_{i=1}^{N} | err_{\text{est}}^{(i)} - err_{\text{ideal}}^{(i)} |}{\sum_{i=1}^{N} err_{\text{ideal}}^{(i)}},
\end{equation}
\noindent where $err_{\text{est}}^{(i)}$ and $err_{\text{ideal}}^{(i)}$ represent the prediction errors (i.e., ADE and FDE) at the $i$-th rank position for the estimated and ideal rankings, respectively.

The results are presented in \cref{TAB:rank_difference}. For the ADE metric, the discrepancy is below 10\% for both Human3.6M and AMASS. For the FDE metric, the discrepancy is slightly higher, at 13.7\% for Human3.6M and 14.4\% for AMASS. The minimal magnitude of these deviations indicates that our uncertainty estimates are highly reliable and exhibit strong correlation with the true prediction error.

\my{
For comparison, we report the results of two uncertainty-aware methods in \cref{TAB:rank_difference}: (1) Motron~\cite{salzmann2022motron}, a parametric approach that provides likelihood-based uncertainty estimates, and (2) ProbHMI~\cite{ma2025probHMI}, a SOTA method specifically designed for uncertainty quantification via latent quantiles. Our results demonstrate a 25.6\% and 26.5\% reduction on Human3.6M and AMASS compared to Motron, while achieving values very similar to ProbHMI. These findings support our claim that the flow-based likelihood produces competitive and well-calibrated uncertainty estimates.
}


\begin{table}[htb]
    \renewcommand{\arraystretch}{1.25}
    \centering
    \caption{The area discrepancy in the error distribution.}
    \resizebox{0.85\linewidth}{!}{
        \begin{tabular}{c|cc|cc}
        \toprule
        \multirow{2}{*}{\diagbox{Method}{Metric}} & \multicolumn{2}{c|}{\textbf{Human3.6M}} & \multicolumn{2}{c}{\textbf{AMASS}} \\
        & ADE & FDE & ADE & FDE \\
        \hline
        Ours & \textbf{0.096} & \textbf{0.137} & \underline{0.083} & \underline{0.144} \\
        Motron~\cite{salzmann2022motron} & 0.129 & 0.151 &  0.113 & 0.152 \\
        ProbHMI~\cite{ma2025probHMI} & \underline{0.109} & \underline{0.139} & \textbf{0.076} & \textbf{0.141} \\
        \bottomrule
        \end{tabular}
    }
    \label{TAB:rank_difference}
\end{table}

\section{\my{Additional Visualization Results}} \label{SEC:additional_visualization}
We present additional comparisons against several recent SOTA methods, including CoMusion~\cite{sun2024comusion} and SLD~\cite{xu2024learning} on Human3.6M, and CoMusion~\cite{sun2024comusion} and SkeletonDiff~\cite{curreli2025nonisotropic} on AMASS, all of which demonstrate strong performance in terms of \textit{accuracy}, \textit{diversity} and \textit{plausibility}. As shown in \cref{fig:qualitative_results}, the first row presents results on Human3.6M, while the second row illustrates results on AMASS, with each frame visualizing 10 stochastic forecasts. 

Although baseline methods can produce diverse motion sequences, our approach generates more natural and realistic pose forecasts, exhibiting higher \textit{plausibility} than the baselines. For example, on AMASS, CoMusion and SkeletonDiff often produce poses with unnaturally lifted legs, as highlighted in the left (CoMusion) and right (SkeletonDiff) yellow boxes, respectively. Moreover, motions generated by the baselines frequently deviate from the contextual cues. As highlighted in the middle green box on Human3.6M, both CoMusion and SLD tend to produce poses with a bent upper body, which is inconsistent with the intended walking motion. Similarly, in the yellow boxes on AMASS, some baseline-generated motions depict unnatural sitting postures. Another example of implausible predictions from the baselines is their tendency to generate poses with elevated body parts, such as arms positioned unnaturally far from the torso, as illustrated in the left and right green boxes and the middle yellow box. We argue that the inferior \textit{plausibility} of baseline methods stems from their inherently single-modal assumptions for motion modeling, which can inadvertently fuse incompatible semantics (e.g., walking vs. sitting). In contrast, our method proposes a multi-modal prior that effectively disentangles these semantic mixtures, thereby producing motions that are both natural and semantically consistent.

\begin{figure*}[htb]
  \centering
  \resizebox{1.0\linewidth}{!}{
    \begin{tabular}{cc} 
        \vspace{5.0em}
        
        \begin{tabular}{c} \huge
            \rotatebox[origin=c]{90}{Human3.6M}
        \end{tabular} &
        \begin{tabular}{cccc}
            \begin{tabular}{l}
                \subfloat{\includegraphics[width=1.0\linewidth]{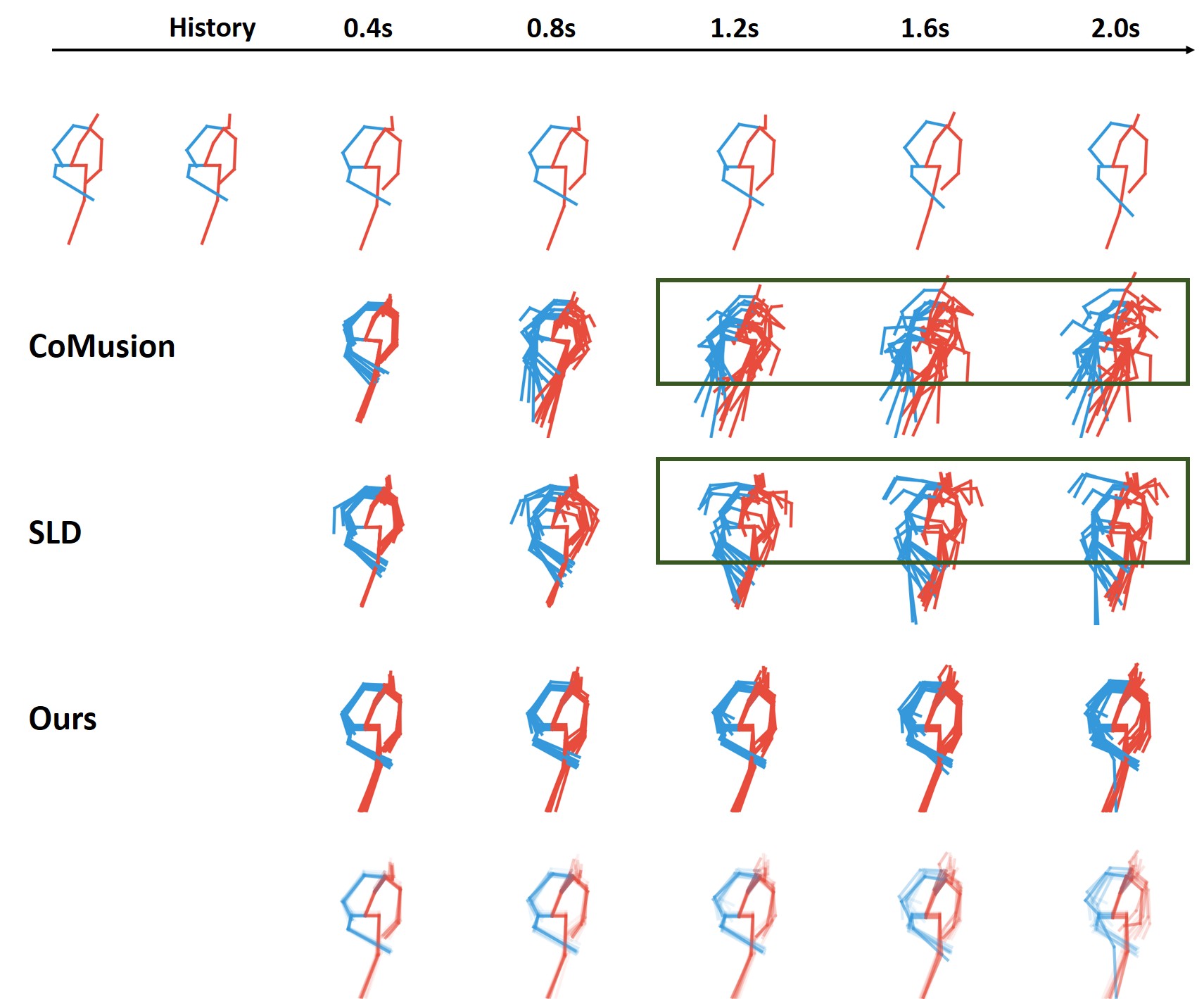}} 
            \end{tabular} &
            
            \begin{tabular}{l}
                \subfloat{\includegraphics[width=1.0\linewidth]{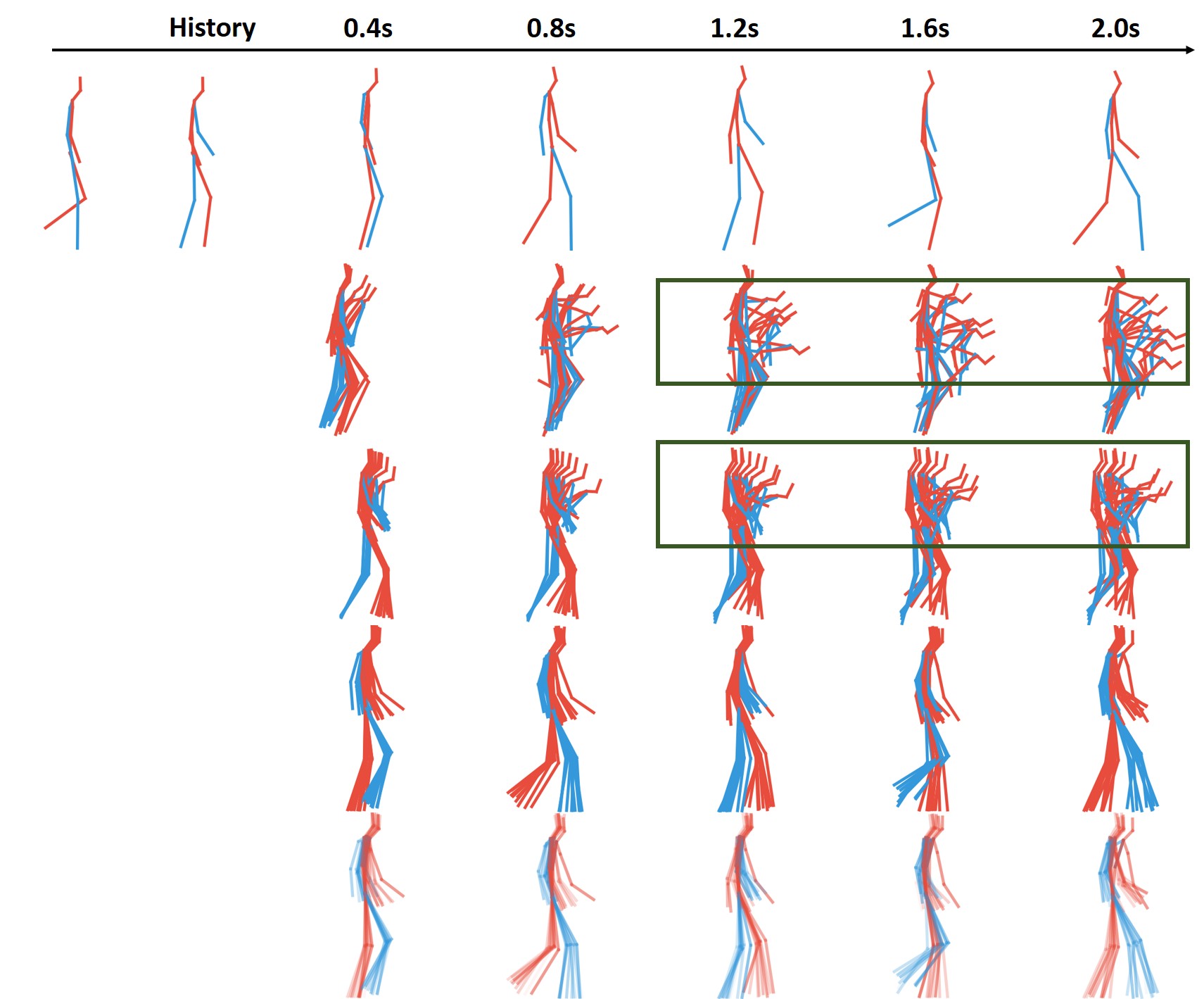}} 
            \end{tabular} &
    
            \begin{tabular}{l}
                \subfloat{\includegraphics[width=1.0\linewidth]{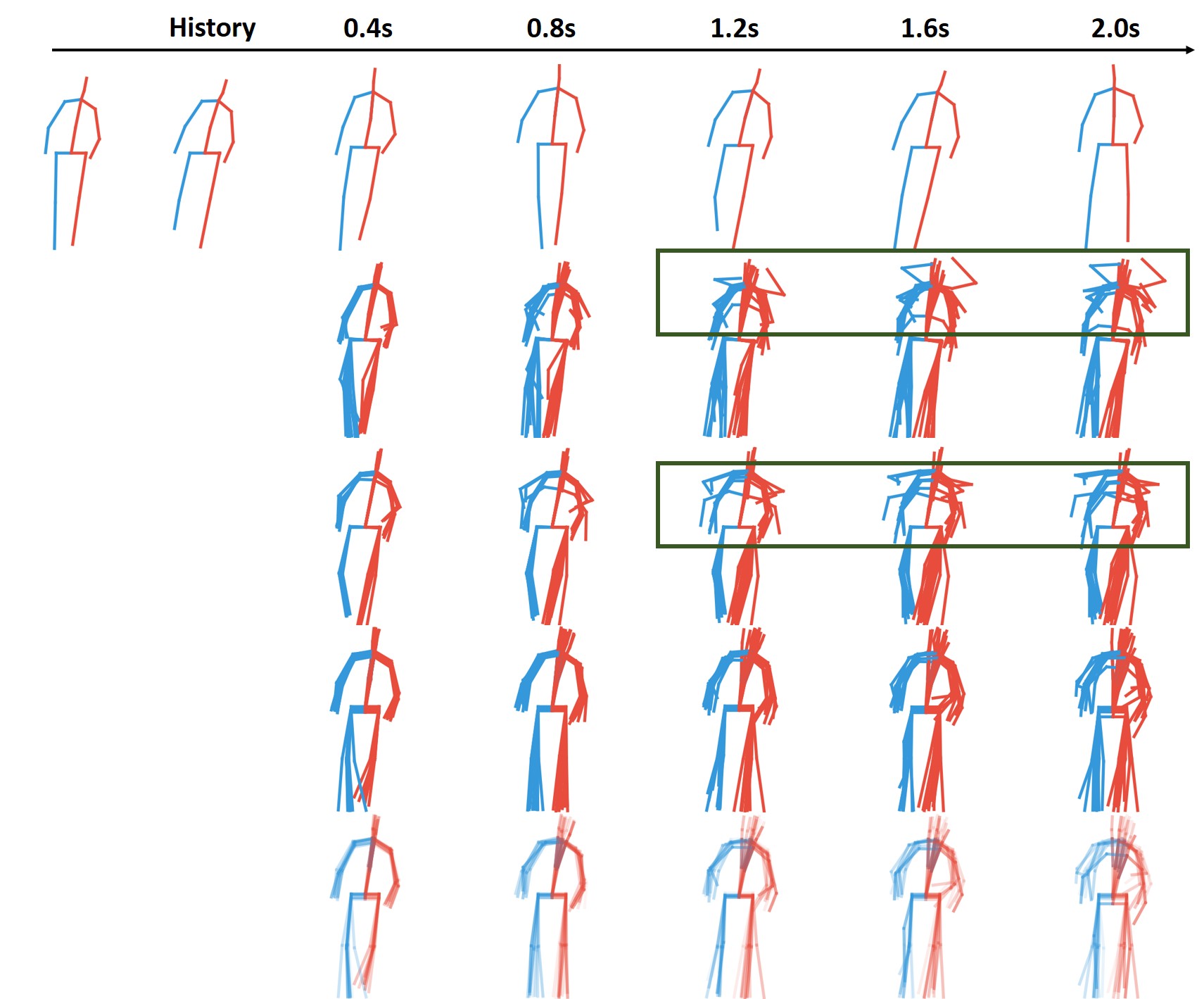}} 
            \end{tabular} 
        \end{tabular} \\
        
        \begin{tabular}{c} \huge
            \rotatebox[origin=c]{90}{AMASS}
        \end{tabular} &
        \begin{tabular}{cccc}
            \begin{tabular}{l}
                \subfloat{\includegraphics[width=1.0\linewidth]{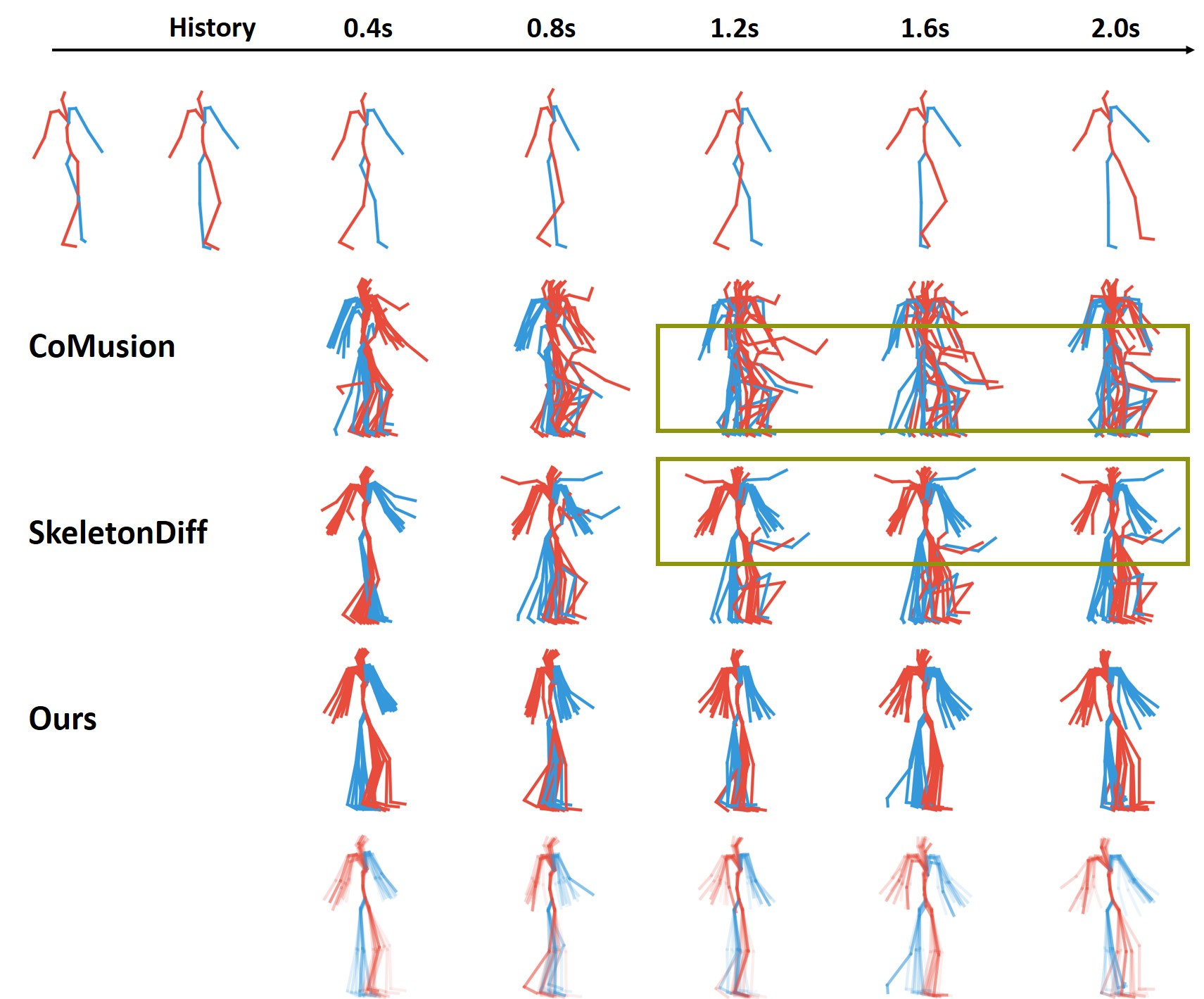}} 
            \end{tabular} &
            
            \begin{tabular}{l}
                \subfloat{\includegraphics[width=1.0\linewidth]{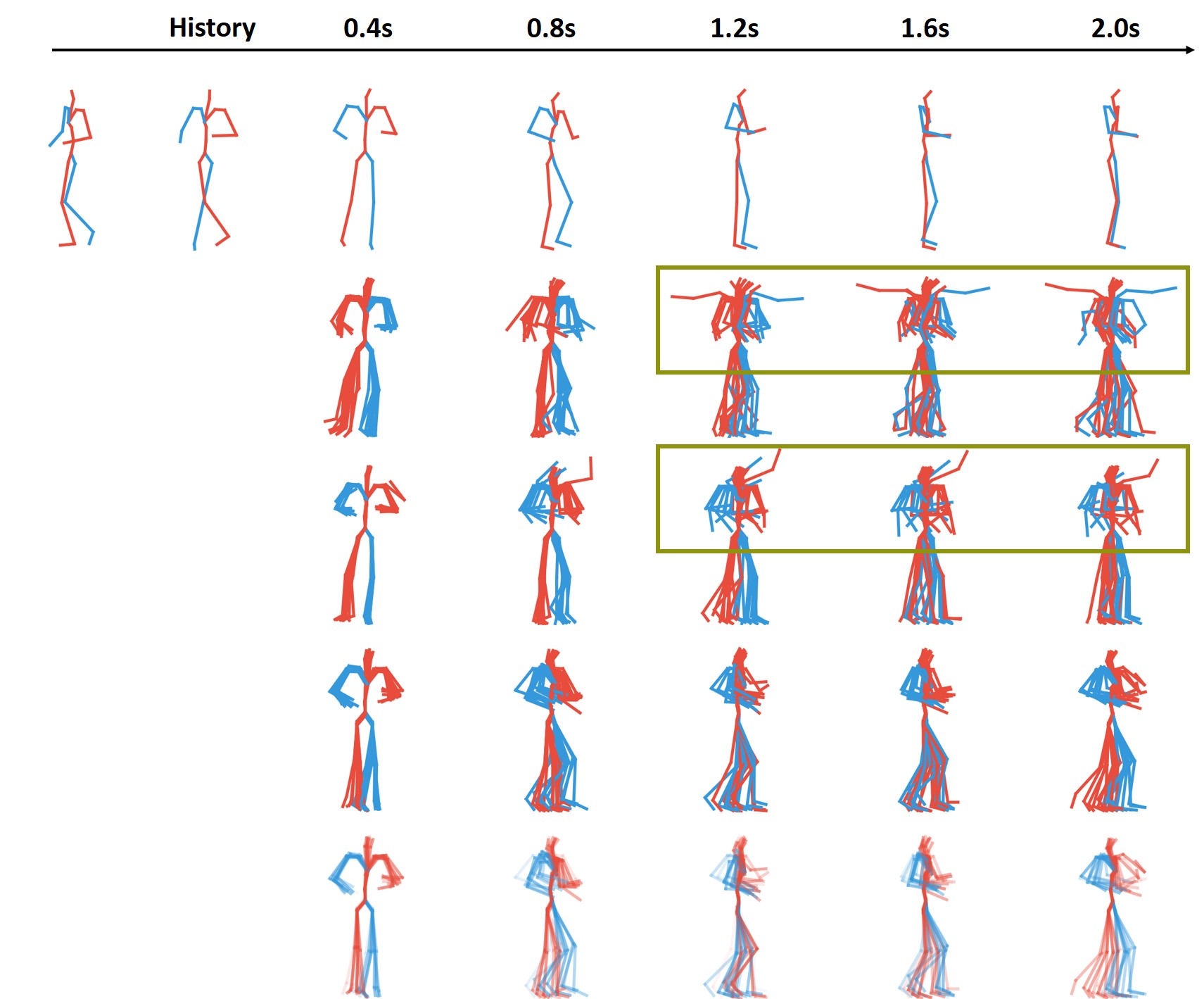}} 
            \end{tabular} &
    
            \begin{tabular}{l}
                \subfloat{\includegraphics[width=1.0\linewidth]{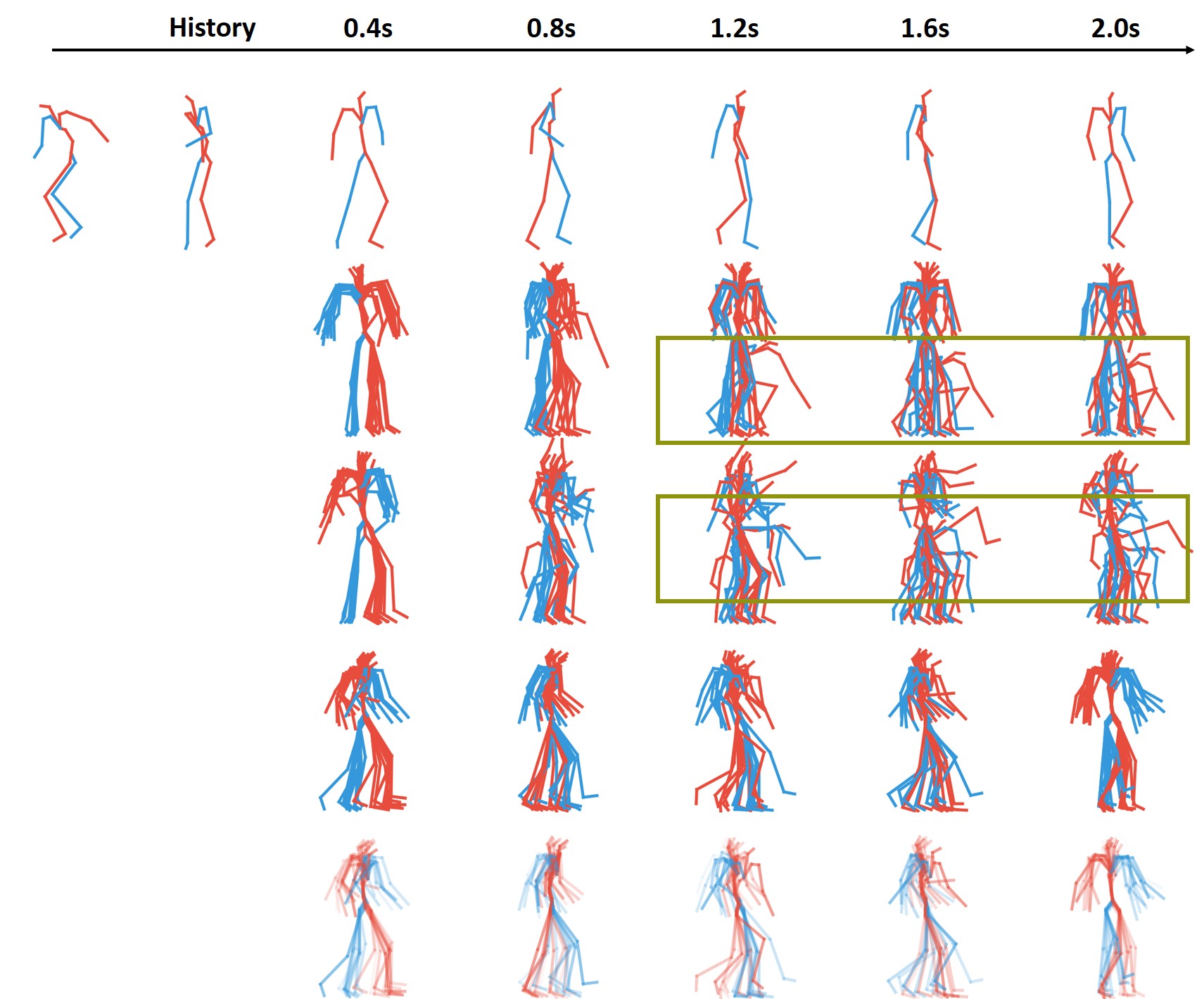}} 
            \end{tabular} 
        \end{tabular}

    \end{tabular}
  }
\caption{Qualitative results. We present qualitative comparison results with CoMusion ~\cite{sun2024comusion} and SLD~\cite{xu2024learning} on Human3.6M (top), and with CoMusion~\cite{sun2024comusion} and SkeletonDiff~\cite{curreli2025nonisotropic} on AMASS (bottom). In each group, the final row shows visualizations weighted by uncertainty as estimated by our model, with greater opacity indicating higher probability.} 
\label{fig:qualitative_results}
\end{figure*}

Moreover, we present qualitative results of uncertainty quantification in the final row of \cref{fig:qualitative_results}, where each visualization is weighted by estimated probability density. These results demonstrate that our method effectively represents future motion distributions with a compact spread, as the high-opacity regions in the density map are concentrated around the ground truth, while lower-opacity regions correspond to predictions with greater deviation. This validates that the estimated probability serves as a reliable measure of uncertainty, enabling the model to differentiate between predictions and prioritize more likely future outcomes, which is crucial for safety-aware applications.

\section{\my{Training and Implementation Details}} \label{SEC:training_details}
Similar to latent diffusion-based methods~\cite{barquero2023belfusion,ho2020denoising}, we adopt a two-stage training schedule. In the first stage, we train the flow model using the EM algorithm described in \cref{SEC:latent_space}, optimized with AdamW at a learning rate of $2\times10^{-4}$ for 20 epochs on both datasets. The batch size is set to 64 for Human3.6M and 128 for AMASS. Since the E-step and M-step for updating the latent distribution parameters are computationally expensive, we perform them once per epoch. Owing to the soft alignment strategy employed for learning the mixture distribution, which can prune components~\cite{dilokthanakul2016deep}, we find that using 16 components is sufficient for both datasets. For training stability, the cluster centers are initialized via latent space clustering, leveraging the model parameters after their initial configuration. In the second stage, we train the motion prediction transformer while keeping the parameters of the flow model frozen. The model is optimized using AdamW with a learning rate of $1\times10^{-3}$ for 30 epochs on both datasets. The batch size is set to 64 for Human3.6M and 128 for AMASS.

We implement a 6-layer flow model as the latent backbone, where the latent dimension at each partition level is set to twice that of the input feature. For Human3.6M, the motion prediction transformer consists of 12 blocks with 6 attention heads per self-attention layer, while for AMASS, it comprises 16 blocks with 8 attention heads. The latent dimension of the transformer is set to 72 for Human3.6M and 256 for AMASS. For both datasets, the number of DCT coefficients $L$ is set to 16. Training details are provided in \appendixref{SEC:training_details}.

\section{Comparison with Deterministic HMP}
By replacing the stochastic initialization $\hat{\mathbf{Z}}_{0} \sim p_{\mathbf{z}_{0}}$ in \cref{EQ:fm_objective} with a deterministic $\mathbf{Z}_{0}$, our framework can be reduced to a deterministic HMP formulation. As the application of flow matching to learn deterministic mappings between paired data has been rarely explored, particularly for high-dimensional regression problems, we compare our approach against state-of-the-art deterministic prediction baselines on the Human3.6M dataset to validate both its performance and that of our overall framework. Since human poses are represented using the exponential map parameterization to preserve bone lengths in our experiments, we adopt the \textbf{Mean Angle Error} as the evaluation metric, which computes the average L2 distance across all joint angles between the deterministic prediction and the ground truth. Following prior studies~\cite{martinez2017human,li2020dynamic,mao2020history,dai2023kd}, we utilize 10 observed frames (0.4s) followed by 25 frames (1s) with a 22-joint skeleton at 25 fps. The training set consists of subjects S1, S7, S6, S8, S9, and S11, with testing on subject S5. We adopt the motion prediction transformer consisting of 16 layers, each equipped with 8 self-attention heads, and set the latent dimension to 128.

The results are summarized in \cref{TAB:short_term_prediction} and \cref{TAB:long_term_prediction}, where the former reports comparisons with baselines for short-term prediction ($\leq$ 400 ms), and the latter presents results for long-term prediction ($\geq$ 560 ms). As shown in \cref{TAB:short_term_prediction}, our method achieves performance comparable to the baselines in short-term prediction and attains the best results on \textit{Smoking}, \textit{Sitting} and \textit{WalkingDog}. For the long-term prediction, our method demonstrates even greater advantages. Specifically, it achieves the best performance on 7 out of 15 action classes, whereas competing methods achieve at most 4. Interestingly, the strengths of our method and the baselines differ notably: while the baselines perform better on \textit{Posing} and \textit{Directions}, our method achieves substantially superior performance on \textit{WalkingDog} and \textit{Smoking}. We also report the results of our method without the latent space (denoted as w/ FM only). As shown in \cref{TAB:short_term_prediction} and \cref{TAB:long_term_prediction}, incorporating the latent space yields only marginal improvements, indicating that this latent representation provides limited benefits for deterministic motion prediction, which fundamentally differ from stochastic prediction.
 
\begin{table*}[htb]
    \renewcommand{\arraystretch}{1.25}
    \centering
    \caption{The results of short-term prediction ($\le$ 400ms) compared to baselines on Human3.6M. The best results are highlighted in \textbf{bold}.}
    \resizebox{1.0\linewidth}{!}{
        \begin{threeparttable}
        \begin{tabular}{l|cccc|cccc|cccc|cccc}
        \toprule
            & \multicolumn{4}{c}{Walking} 
            & \multicolumn{4}{c}{Eating} 
            & \multicolumn{4}{c}{Smoking}
            & \multicolumn{4}{c}{Discussion} \\
            Milliseconds & 80 & 160 & 320 & 400 & 80 & 160 & 320 & 400 & 80 & 160 & 320 & 400 & 80 & 160 & 320 & 400 \\
        \hline
            ResGRU~\cite{martinez2017human} & 0.28 & 0.49 & 0.72 & 0.81 & 0.23 & 0.39 & 0.62 & 0.76 & 0.23 & 0.39 & 0.62 & 0.76 & 0.31 & 0.68 & 1.01 & 1.09 \\
            DMGNN~\cite{li2020dynamic} & 0.18 & 0.31 & 0.49 & 0.58 & 0.17 & 0.30 & \textbf{0.49} & 0.59 & 0.21 & 0.39 & 0.81 & 0.77 & 0.26 & 0.65 & 0.92 & 0.99 \\
            Hisrep~\cite{mao2020history} & 0.18 & \textbf{0.30} & \textbf{0.46} & \textbf{0.51} & 0.16 & 0.29 
            & \textbf{0.49} & 0.60 & 0.22 & 0.40 & 0.86 
            & 0.80 & 0.20 & \textbf{0.52} & \textbf{0.78} & 0.87 \\
            KD-Former~\cite{dai2023kd} & \textbf{0.15} & 0.32 & 0.54 & 0.61 & \textbf{0.14} & 0.28 & 0.50 & \textbf{0.51} & 0.17 & 0.37 & 0.76 & 1.46 & 0.19 & 0.53 & 0.87 & 0.90 \\
            MSTP-Net~\cite{chen2023mstp} & \textbf{0.19} & 0.34 & 0.50 & 0.54 & 0.16 & 0.29
            & 0.50 & 0.61 & 0.21 & 0.40 & 0.80 
            & 0.78 & 0.21 & 0.54 & 0.79 & \textbf{0.83} \\
        \hline
            Ours w/ FM only & 0.37 & 0.47 & 0.68 & 0.77 & 0.15 & \textbf{0.22} & \textbf{0.49} & 0.57 & 0.12 	& \textbf{0.22} & 0.35 & 0.42 & 0.47 & 0.93
            & 1.09 & 1.05 \\ 
            Ours & 0.36 & 0.46 & 0.73 & 0.84 & \textbf{0.14} & 0.24 & 0.51 & 0.58 & \textbf{0.11} & \textbf{0.22} & \textbf{0.34} & \textbf{0.40} & 0.47 & 0.93 & 1.11 & 1.05 \\ 

        \midrule
            & \multicolumn{4}{c}{Direction} 
            & \multicolumn{4}{c}{Greeting} 
            & \multicolumn{4}{c}{Phoning}
            & \multicolumn{4}{c}{Posing} \\
            Milliseconds & 80 & 160 & 320 & 400 & 80 & 160 & 320 & 400 & 80 & 160 & 320 & 400 & 80 & 160 & 320 & 400 \\
        \hline
            ResGRU~\cite{martinez2017human} & 0.26 & 0.47 & 0.72 & 0.84 & 0.75 &  1.17 & 1.74 & 1.83 & \textbf{0.23} & 0.43 & \textbf{0.69} & \textbf{0.82} & 0.36 & 0.71 & 1.22 & 1.48 \\
            DMGNN~\cite{li2020dynamic} & 0.32 & 0.65 & 0.93 & 1.05 & 0.36 & 0.61 & \textbf{0.94} & \textbf{1.12} & 0.52 & 0.97 & 1.29 & 1.43 & 0.20 & 0.46 & 1.06 & 1.34 \\
            Hisrep~\cite{mao2020history} & 0.25 & 0.43 & \textbf{0.60} & \textbf{0.69} & 0.35 & 0.60 
            & 0.95 & 1.14 & 0.53 & 1.01 & 1.31 
            & 1.43 & 0.19 & 0.46 & 1.09 & 1.35 \\
            KD-Former~\cite{dai2023kd} & \textbf{0.24} & 0.52 & 0.72 & 0.77 & \textbf{0.27} & 0.72 & 1.11 & 1.25 & \textbf{0.17} & 0.66 & 1.28 & 1.35 & \textbf{0.17} & \textbf{0.43} & \textbf{0.92} & \textbf{1.18} \\
            MSTP-Net~\cite{chen2023mstp} & 0.27 & \textbf{0.42} & 0.63 & \textbf{0.69} & 0.36 & 0.64 
            & 1.01 & 1.17 & 0.50 & 0.99 & 1.32 
            & 1.46 & 0.20 & 0.50 & 1.10 & 1.33 \\
        \hline
            Ours w/ FM only & 0.26 & 0.56 & 1.00 & 1.14 & 0.40 & \textbf{0.58} & 1.23 & 1.37 
            & 0.30 & 0.46 & 0.80 & 0.96 & 0.33
            & 0.65 & 1.17 & 1.49 \\ 
            Ours & \textbf{0.24} & 0.53 & 0.95 & 1.07 & 0.36 & 0.60 & 1.08 & 1.32 & 0.26 & \textbf{0.42} & 0.81 & 0.98 & 0.29 & 0.60 & 1.16 & 1.50  \\

        \midrule
            & \multicolumn{4}{c}{Purchase} 
            & \multicolumn{4}{c}{Sitting} 
            & \multicolumn{4}{c}{SittingDown}
            & \multicolumn{4}{c}{TakingPhoto} \\
            Milliseconds & 80 & 160 & 320 & 400 & 80 & 160 & 320 & 400 & 80 & 160 & 320 & 400 & 80 & 160 & 320 & 400 \\
        \hline
            ResGRU~\cite{martinez2017human} & 0.51 & 0.97 & 1.07 & 1.16 & 0.41 & 1.05 & 1.49 & 1.63 & 0.39 & 0.81 & 1.40 & 1.62 & 0.24 & 0.51 & 0.90 & 1.05 \\
            DMGNN~\cite{li2020dynamic} & 0.41 & \textbf{0.61} & 1.05 & 1.14 & 0.26 & 0.42 & 0.76 & 0.97 & 0.32 & 0.65 & 0.93 & 1.05 & 0.15 & \textbf{0.34} & 0.58 & 0.71 \\
            Hisrep~\cite{mao2020history} & 0.42 & 0.65 & 1.00 & 1.07 & 0.29 & 0.47 & 0.83 & 1.01 & 0.30 & 0.63 & 0.92 & 1.04 & 0.16 & 0.36 & 0.58 & 0.70 \\
            KD-Former~\cite{dai2023kd} & \textbf{0.26} & 0.72 & 0.97 & 1.07 & 0.23 & 0.53 & 0.94 & 1.61 & \textbf{0.26} & 0.63 & 0.98 & 1.12 & \textbf{0.15} & 0.39 & 0.72 & 0.84 \\
            MSTP-Net~\cite{chen2023mstp} & 0.47 & 0.68 & 1.00 & \textbf{1.06} & 0.29 & 0.45 
            & 0.81 & 0.99 & 0.30 & \textbf{0.62} & \textbf{0.86} 
            & \textbf{0.96} & 0.16 & 0.37 & 0.60 & 0.71 \\
        \hline
            Ours w/ FM only & 0.46 & 0.72 & 0.94 & 1.18 & \textbf{0.17} & 0.39 & 0.74 & 0.97 
            & 0.33 & 0.72 & 1.16 & 1.29 & 0.25 
            & \textbf{0.34} & \textbf{0.56} & 0.69 \\ 
            Ours & 0.45 & 0.75 & \textbf{0.89} & 1.13 & \textbf{0.17} & \textbf{0.38} & \textbf{0.72} & \textbf{0.95} & 0.37 & 0.79 & 1.26 & 1.38 & 0.26 & 0.37 & \textbf{0.56} & \textbf{0.67} \\

        \midrule
            & \multicolumn{4}{c}{Waiting} 
            & \multicolumn{4}{c}{WalkingDog} 
            & \multicolumn{4}{c}{WalkingTogether}
            & \multicolumn{4}{c}{Average} \\
            Milliseconds & 80 & 160 & 320 & 400 & 80 & 160 & 320 & 400 & 80 & 160 & 320 & 400 & 80 & 160 & 320 & 400 \\
        \hline
            ResGRU~\cite{martinez2017human} & 0.28 & 0.53 & 1.02 & 1.14 & 0.56 & 0.91 & 1.26 & 1.40 & 0.31 & 0.58 & 0.87 & 0.91 & 0.36 & 0.67 & 1.02 & 1.15 \\
            DMGNN~\cite{li2020dynamic} & 0.22 & 0.49 & \textbf{0.88} & \textbf{1.10} & 0.42 & 0.72 & 1.16 & 1.34 & 0.15 & 0.33 & \textbf{0.50} & 0.57 & 0.27 & 0.52 & 0.83 & 0.95 \\
            Hisrep~\cite{mao2020history} & 0.22 & 0.49 & 0.92 & 1.14 & 0.46 & 0.78 & 1.05 & 1.23 & \textbf{0.14} & \textbf{0.32} & \textbf{0.50} & \textbf{0.55} & 0.27 & 0.52 & \textbf{0.82} & 0.94 \\
            KD-Former~\cite{dai2023kd} & \textbf{0.18} & \textbf{0.47} & 0.98 & 1.15 & 0.31 & 0.74 & 1.12 & 1.35 & 0.15 & 0.39 & 0.55 & 0.62 & \textbf{0.20} & \textbf{0.51} & 0.86 & 1.01 \\
            MSTP-Net~\cite{chen2023mstp} & 0.23 & 0.50 & 0.92 & 1.12 & 0.47 & 0.78 & 1.08 & 1.21 & 0.17 & 0.38 & 0.53 & 0.57 & 0.28 & 0.53 & 0.83 & \textbf{0.93} \\
        \hline
            Ours w/ FM only & 0.22 & 0.75 & 1.06 & 1.26 & 0.32 & 0.52 & 0.91 & 1.16 & 0.27 & 0.48 & 0.64 & 0.68 & 0.29 & 0.53 & 0.85 & 1.00 \\ 
            Ours & 0.22 & 0.74 & 1.03 & 1.22 & \textbf{0.30} & \textbf{0.51} & \textbf{0.89} & \textbf{1.13} & 0.24 & 0.42 & 0.55 & 0.57 & 0.28 & 0.53 & 0.84 & 0.99 \\
        \bottomrule 
        \end{tabular}
        \end{threeparttable}
    }
    \label{TAB:short_term_prediction}
\end{table*}

\begin{table*}[htb]
    \renewcommand{\arraystretch}{1.25}
    \centering
    \caption{The results of long-term prediction ($\ge$ 560ms) compared to baselines on Human3.6M. The best results are highlighted in \textbf{bold}.}
    \resizebox{1.0\linewidth}{!}{
        \begin{threeparttable}
        \begin{tabular}{l|cccc|cccc|cccc|cccc}
        \toprule
            & \multicolumn{4}{c}{Walking} 
            & \multicolumn{4}{c}{Eating} 
            & \multicolumn{4}{c}{Smoking}
            & \multicolumn{4}{c}{Discussion} \\
            Milliseconds & 560 & 720 & 880 & 1000 & 560 & 720 & 880 & 1000 & 560 & 720 & 880 & 1000 & 560 & 720 & 880 & 1000 \\
        \hline
            ResGRU~\cite{martinez2017human} & 0.93 & - & - & 1.03 & 0.95 & - & - & 1.08 & 1.25 & - & - & 1.50 & 1.43 & - & - & 1.69 \\
            DMGNN~\cite{li2020dynamic} & \textbf{0.66} & - & - & 0.75 & 0.74 & - & - & 1.14 & 0.83 & - & - & 1.52 & 1.33 & - & - & 1.45 \\
            Hisrep~\cite{mao2020history} & 0.59 & \textbf{0.62} & \textbf{0.61} & 0.64 & 0.74 & 0.81 & 1.01 & 1.10 & 0.86 & 1.00 & 1.35 & 1.58 & 1.29 & 1.51 & 1.66 & 1.63 \\
            KD-Former~\cite{dai2023kd} & 0.70 & - & - & 0.69 & \textbf{0.71} & - & - & \textbf{1.08} & 1.01 & - & - & 1.46 & 1.24 & - & - & 1.69 \\
            MSTP-Net~\cite{chen2023mstp} & 0.60 & 0.66 & 0.67 & 0.68 & 0.72 & \textbf{0.78} & \textbf{0.96} & \textbf{1.08} & 0.86 & 0.98 & 1.30 & 1.51 & 1.22 & 1.42 & 1.49 & 1.51 \\
        \hline
            Ours w/ FM only & 0.86 & 0.84 & 0.84 & 0.87 & 0.83 & 1.05 & 1.19 & 1.36 
            & 0.61 & 0.82 & \textbf{1.00} & 1.07 & 1.24 
            & 1.48 & 1.45 & 1.53 \\ 
            Ours & 0.91 & 0.89 & 0.86 & 0.89 & 0.82 & 1.03 & 1.18 & 1.35 & \textbf{0.60} & \textbf{0.81} & \textbf{1.00} & \textbf{1.08} & \textbf{1.21} & \textbf{1.41} & \textbf{1.40} & \textbf{1.48} \\  

        \midrule
            & \multicolumn{4}{c}{Direction} 
            & \multicolumn{4}{c}{Greeting} 
            & \multicolumn{4}{c}{Phoning}
            & \multicolumn{4}{c}{Posing} \\
            Milliseconds & 560 & 720 & 880 & 1000 & 560 & 720 & 880 & 1000 & 560 & 720 & 880 & 1000 & 560 & 720 & 880 & 1000 \\
        \hline
            ResGRU~\cite{martinez2017human} & 1.15 & - & - & 1.64 & 1.82 & - & - & 2.14 & 1.55 & - & - & 2.05 & 2.39 & - & - & 2.89 \\
            DMGNN~\cite{li2020dynamic} & 0.86 & - & - & 1.30 & 1.57 & - & - & 1.63 & 1.44 & - & - & 1.64 & 1.49 & - & - & \textbf{2.17} \\
            Hisrep~\cite{mao2020history} & 0.81 & 1.02 & 1.22 & 1.27 & 1.47 & 1.47 & 1.61 & 1.57 & 1.41 & 1.55 & 1.68 & 1.68 & 1.60 & \textbf{1.78} & \textbf{2.10} & 2.32 \\
            KD-Former~\cite{dai2023kd} & 0.88 & - & - & 1.36 & 1.53 & - & - & 1.89 & 1.54 & - & - & 1.95 & \textbf{1.53} & - & - & 2.29 \\
            MSTP-Net~\cite{chen2023mstp} & \textbf{0.78} & \textbf{0.95} & \textbf{1.16} & \textbf{1.17} & 1.44 & 1.41 & 1.56 & 1.51 & \textbf{1.38} & \textbf{1.48} & \textbf{1.56} & \textbf{1.54} & 1.54 & 1.83 & 2.14 & 2.29 \\
        \hline
            Ours w/ FM only & 1.29 & 1.43 & 1.48 & 1.59 & \textbf{1.32} & \textbf{1.30} & \textbf{1.22} & \textbf{1.37} 
            & 1.47 & 1.65 & 1.67 & 1.80 & 2.12 	& 2.32 & 2.51 & 2.64 \\  
            Ours & 1.20 & 1.37 & 1.43 & 1.57 & \textbf{1.32} & 1.34 & 1.25 & 1.42 & 1.43 & 1.61 & 1.62 & 1.76 & 2.15 & 2.36 & 2.56 & 2.68 \\ 

        \midrule
            & \multicolumn{4}{c}{Purchase} 
            & \multicolumn{4}{c}{Sitting} 
            & \multicolumn{4}{c}{SittingDown}
            & \multicolumn{4}{c}{TakingPhoto} \\
            Milliseconds & 560 & 720 & 880 & 1000 & 560 & 720 & 880 & 1000 & 560 & 720 & 880 & 1000 & 560 & 720 & 880 & 1000 \\
        \hline
            ResGRU~\cite{martinez2017human} & 1.45 & - & - & 2.35 & 1.66 & - & - & 1.91 & 1.40 & - & - & 2.06 & 0.88 & - & - & 1.10 \\
            DMGNN~\cite{li2020dynamic} & 1.39 & - & - & 2.13 & \textbf{1.12} & - & - & \textbf{1.51} & 1.30 & - & - & 1.74 & 0.83 & - & - & 1.06 \\
            Hisrep~\cite{mao2020history} & 1.43 & 1.53 & 1.94 & 2.22 & 1.16 & 1.29 & 1.50 & 1.55 & 1.18 & 1.42 & 1.55 & 1.70 & 0.82 & 0.91 & 1.00 & 1.08 \\
            KD-Former~\cite{dai2023kd} & 1.29 & - & - & 2.13 & 1.71 & - & - & 1.97 & 1.36 & - & - & 1.90 & 1.00 & - & - & 1.26 \\
            MSTP-Net~\cite{chen2023mstp} & 1.37 & 1.42 & 1.88 & 2.18 & 1.13 & \textbf{1.26} & 1.48 & 1.54 & \textbf{1.08} & \textbf{1.30} & \textbf{1.44} & \textbf{1.60} & \textbf{0.79} & \textbf{0.85} & \textbf{0.91} & \textbf{0.99} \\
        \hline
            Ours w/ FM only & \textbf{1.28} & 1.20 & \textbf{1.32} & \textbf{1.35} & 1.25 & 1.32 & 1.47 & 1.64 
            & 1.48 & 1.69 & 1.84 & 1.95 & 0.99 	& 1.22 & 1.41 & 1.59 \\
            Ours & \textbf{1.28} & \textbf{1.15} & 1.33 & 1.40 & 1.23 & \textbf{1.26} & \textbf{1.41} & 1.57 & 1.54 & 1.70 & 1.84 & 1.96 & 0.97 & 1.17 & 1.37 & 1.55 \\ 

        \midrule
            & \multicolumn{4}{c}{Waiting} 
            & \multicolumn{4}{c}{WalkingDog} 
            & \multicolumn{4}{c}{WalkingTogether}
            & \multicolumn{4}{c}{Average} \\
            Milliseconds & 560 & 720 & 880 & 1000 & 560 & 720 & 880 & 1000 & 560 & 720 & 880 & 1000 & 560 & 720 & 880 & 1000 \\
        \hline
            ResGRU~\cite{martinez2017human} & 1.64 & - & - & 2.22 & 1.66 & - & - & 1.92 & 1.14 & - & - & 1.61 & 1.57 & - & - & 2.04 \\
            DMGNN~\cite{li2020dynamic} & 1.46 & - & - & 2.12 & 1.57 & - & - & 1.75 & 0.70 & - & - & 1.24 & 1.17 & - & - & 1.57 \\
            Hisrep~\cite{mao2020history} & 1.54 & 1.90 & 2.22 & 2.30 & 1.57 & 1.63 & 1.76 & 1.82 & 0.63 & 0.68 & 0.79 & 1.16 & 1.14 & 1.28 & 1.46 & 1.57 \\ 
            KD-Former~\cite{dai2023kd} & 1.50 & - & - & 2.35 & 1.48 & - & - & 1.79 & 0.68 & - & - & 1.11 & 1.21 & - & - & 1.66 \\
            MSTP-Net~\cite{chen2023mstp} & 1.46 & 1.80 & 2.13 & 2.18 & 1.49 & 1.55 & 1.69 & 1.75 & 0.61 & 0.72 & 0.81 & 1.14 & \textbf{1.10} & \textbf{1.23} & 1.41 & \textbf{1.51} \\ 
        \hline
            Ours w/ FM only & \textbf{1.42} & \textbf{1.56} & \textbf{1.74} & \textbf{1.75} & 1.29 & 1.40 & 1.32 & 1.36 
            & 0.75 & 0.75 & 0.80 & 0.85 & 1.21	& 1.33 & 1.42 &	\textbf{1.51} \\ 
            Ours & \textbf{1.42} & \textbf{1.56} & 1.77 & 1.81 & \textbf{1.28} & \textbf{1.36} & \textbf{1.30} & \textbf{1.34} & \textbf{0.60} & \textbf{0.69} & \textbf{0.80} & \textbf{0.81} & 1.20 & 1.31 & \textbf{1.40} & \textbf{1.51} \\ 
    
        \bottomrule 
        \end{tabular}
        \end{threeparttable}
    }
    \label{TAB:long_term_prediction}
\end{table*}

\end{document}